\documentclass[sigconf,natbib=true,screen=true]{acmart}

\AtBeginDocument{%
  }

\copyrightyear{2026}
\acmYear{2026}
\setcopyright{cc}
\setcctype{by}
\acmConference[CIKM '26]{Proceedings of the 35th ACM International Conference on Information and Knowledge Management}{November 07--11, 2026}{Rome, Italy}
\acmBooktitle{Proceedings of the 35th ACM International Conference on Information and Knowledge Management (CIKM '26), November 07--11, 2026, Rome, Italy}
\acmDOI{10.1145/3799682.3841076}
\acmISBN{979-8-4007-2539-5/2026/11}

\ccsdesc[500]{Computing methodologies~Natural language processing}

\keywords{Fact-Checking, Retrieval-Augmented Generation}

\usepackage{enumitem}
\usepackage{multirow}
\usepackage{balance}
\usepackage[normalem]{ulem}
\useunder{\uline}{\ul}{}
\newcommand{\IS}[1][IS]{#1}
\begin{document}

\title{Evaluating and Improving Evidence-Grounded Fact-Checking in LLMs via Multi-Round Evidence Ablation}


\author{Xingyu Deng}
\authornotemark[1]
\orcid{0009-0007-0832-2572}
\affiliation{%
  \institution{University of Sheffield}
  \city{Sheffield}
  \country{UK}}
\email{xdeng37@sheffield.ac.uk}

\author{Mingzi Cao}
\authornotemark[1]
\orcid{0009-0001-8267-3457}
\affiliation{%
  \institution{University of Sheffield}
  \city{Sheffield}
  \country{UK}}
\email{mcao20@sheffield.ac.uk}

\author{Nikolaos Aletras}
\orcid{0000-0003-4285-1965}
\affiliation{%
  \institution{University of Sheffield}
  \city{Sheffield}
  \country{UK}}
\email{n.aletras@sheffield.ac.uk}

\author{Xi Wang}
\orcid{0000-0001-5936-9919}
\affiliation{%
  \institution{University of Sheffield}
  \city{Sheffield}
  \country{UK}}
\email{xi.wang@sheffield.ac.uk}

\author{Mark Stevenson}
\orcid{0000-0002-9483-6006}
\affiliation{%
  \institution{University of Sheffield}
  \city{Sheffield}
  \country{UK}}
\email{mark.stevenson@sheffield.ac.uk}


\begin{abstract}
Automatic fact-checking systems assess the veracity of claims given evidence from relevant documents. Large Language Models (LLMs) have demonstrated strong performance in fact-checking due to their general reasoning capabilities. However, it remains unclear whether they faithfully make use of the evidence provided to reach veracity judgments or rely on parametric knowledge. To investigate this, we introduce \textbf{F}act-\textbf{A}blated \textbf{E}valuation (\textbf{FAE}), a new evaluation framework that iteratively ablates the cited evidence to assess whether LLMs revise their predictions accordingly. Our empirical results show that current off-the-shelf LLMs as fact-checking systems rely more on their parametric knowledge than on the evidence provided. To bridge this gap between prediction accuracy and evidence grounding, we propose \textbf{REAL} (\textbf{R}igorous \textbf{E}vidence \textbf{A}blation \textbf{L}earning), a training framework that promotes evidence-dependent verification through counterfactual evidence supervision for the LLM-as-verifier models. Experiments on four fact-checking datasets across different domains demonstrate that models trained with REAL obtain superior evidence-dependent capabilities compared to standard fine-tuned models. Our findings highlight that strong fact-checking performance can still coexist with weak evidence dependency, while REAL encourages veracity predictions to remain more closely tied to the availability of supporting evidence.

\end{abstract}

\maketitle

\section{Introduction}
Automated fact checking, the process of assessing the veracity of claims, acts as a safeguard against misinformation \cite{guo2022survey,zeng2021automated}. 
Fact-checking systems are expected not only to predict claim veracity labels, but also to retrieve and present supporting evidence from external knowledge sources such as textual documents and knowledge graphs \cite{guo-etal-2022-survey,zeng2021automated}. 
This evidence helps to explain algorithm decisions, which is particularly important in high-stakes domains, like public health or law, and supports the integration of fact-checking systems into human decision-making workflows \cite{vladika-matthes-2023-scientific}.

The traditional approach to fact-checking involves a two-stage approach: (1) identification of evidence from a collection ranked by retrievers \cite{chen2022gere,wu2021evidence,hanselowski-etal-2019-richly,hu2023read,zheng-etal-2024-evidence,deng2025+,luken-etal-2018-qed,nie2019combining} followed by (2) veracity prediction \cite{vlachos-riedel-2014-fact,guo-etal-2022-survey,zeng2021automated}.
More recently, Large Language Models (LLMs) have demonstrated strong fact-checking performance through application of Retrieval-Augmented Generation (RAG) architectures \cite{lewis2020retrieval} in which retrieved documents are used as additional context prior to veracity prediction \cite{schlichtkrull-etal-2024-automated,akhtar-etal-2025-2nd,pan-etal-2023-fact,lewis2020retrieval}.

However, in RAG settings, LLMs can make use of both retrieved evidence and memorised knowledge when assessing claims, making it difficult to determine the source of their predictions. Since parametric knowledge can become outdated \cite{chenghaozhu2025your} or reflect inaccuracies introduced during training \cite{das2025security}, reliance on such knowledge may affect the reliability of fact-checking decisions \cite{vladika-etal-2025-facts,kassem-etal-2025-alpaca,deng2025next,vladika-matthes-2024-improving, deng2026towards}. 
Strong performance may partly reflect prior exposure to relevant facts included in pretraining data, making it difficult to determine whether predictions are primarily supported by retrieved evidence or memorised knowledge \cite{wei2025memorization,di2025llms,carlini2021extracting}.
Consequently, superior verification accuracy alone cannot reveal whether a verifier model truly depends on the provided evidence or merely recovers memorised knowledge.

\begin{figure}[t]
    \centering
    \includegraphics[width=1\linewidth]{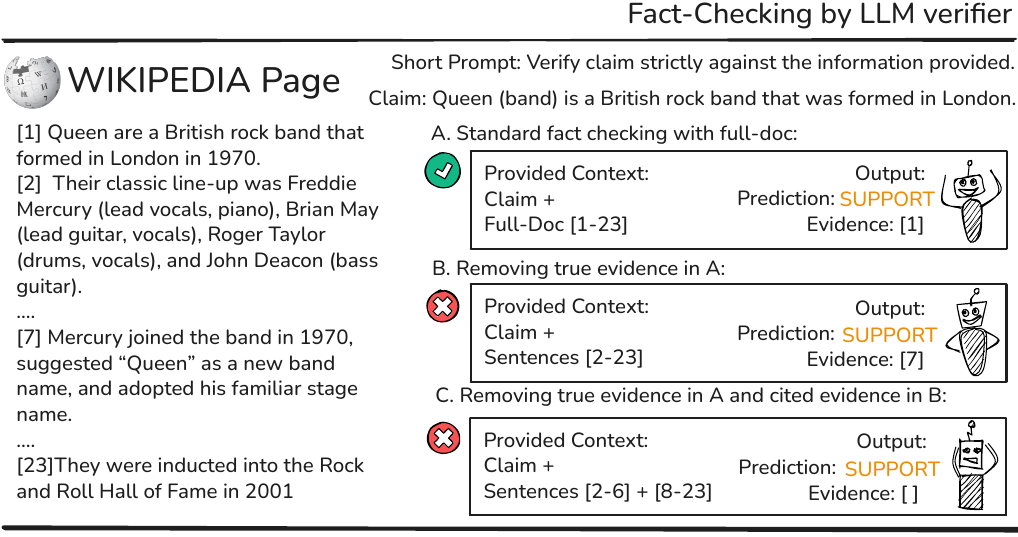}
    \caption{Observed failure mode in FEVER dataset produced by LLM verifier (LLaMa3-8B-Instruct) }
    \Description{The figure about some failure examples observed in FEVER verified by Llama3-8B}
    \vspace{-20pt}
    \label{fig:failure_example}
\end{figure}

To ensure evidence-based fact-checking, verification decisions are expected to be grounded in the evidence cited as their justification. This expectation demands that the decision should be sensitive to changes in the cited evidence, especially when that evidence is no longer available. 
As shown in Figure~\ref{fig:failure_example}, if a claim is supported by deterministic evidence (i.e., Sentence [1]), the verifier should no longer be able to confidently justify the same prediction once that evidence is removed.
In practice, however, LLM-based verifiers often maintain the same decision by ``hallucinating'' alternative justifications (e.g., citing Sentence [7]) or without providing evidence. Standard fact-checking metrics \cite{guo-etal-2022-survey,zeng2021automated,vladika-matthes-2023-scientific,vlachos-riedel-2014-fact} fail to capture this behaviour, as they assess label accuracy only once without testing the causal dependence between evidence and prediction. Therefore, evidence grounding requires counterfactual evaluation by observing whether the decision changes when its supporting evidence is ablated.

To address this limitation, we introduce \textbf{Fact Ablated Evaluation (FAE)}, a behavioural evaluation framework that measures the causal dependency between predictions and model-selected evidence through iterative counterfactual intervention.
FAE evaluates LLM-based verifier grounding behaviour through three complementary metrics that quantify decision shifts across the entire ablation process. For claims initially labelled \texttt{SUPPORT} or \texttt{REFUTE}, an evidence-grounded verifier should immediately abstain (i.e., return \texttt{NOT\_ENOUGH\_INFO}) once the supporting evidence is ablated. However, we observe that current LLM-based verifiers stabilise at a non-zero decision plateau across ablations, indicating persistent reliance on parametric knowledge despite explicit control of evidence completeness. 
These observations suggest that current LLM-based fact-checking systems, despite achieving strong veracity prediction accuracy, do not necessarily learn to make predictions that remain dependent on the provided evidence.

To promote truly evidence-grounded fact-checking, we further propose \textbf{REAL} (\textbf{R}igorous \textbf{E}vidence \textbf{A}blation \textbf{L}earning), a training framework for LLM-based verifiers. 
Rather than relying on off-the-shelf LLMs \cite{rothermel-etal-2024-infact,braun2025defame,ullrich-etal-2024-aic,park-etal-2024-dunamu,jannah-etal-2025-multilingual,li2025use,wang-etal-2024-factcheck,wei2024longform,geng2025m4fc,yoon-etal-2024-hero,mohammadkhani-etal-2024-zero,zhou-etal-2025-gqc,xie-etal-2025-fire,yoon-etal-2025-team,ullrich-drchal-2025-aic,chowdhury-etal-2025-fact5} and standard supervised fine-tuning (SFT) on claim-evidence pairs only \cite{shcharbakova-etal-2025-scale, cheung2023factllama, kumar2025improving, putta-etal-2025-claimcheck, yoon-etal-2024-hero}, REAL introduces counterfactual supervision by training on paired inputs with and without supporting evidence.
It enforces abstention when evidence is ablated, thus creating a causal dependency between evidence and verdict.

\vspace{5pt}
\noindent This paper makes three main contributions:
\begin{itemize}[leftmargin=7pt]
\item We identify a failure mode in LLM-based fact-checking, where predictions remain stable when removing supporting evidence.
\item We propose \textbf{Fact Ablated Evaluation (FAE)}, a behavioural evaluation framework that measures evidence dependence through progressive evidence ablation, and show that parametric knowledge contributes to this failure mode.
\item We introduce \textbf{REAL}, a training framework that uses counterfactual supervision and evidence enhancement to improve evidence grounding without sacrificing fact-checking performance.

\end{itemize}

\begin{figure*}[ht]
    \centering
    \includegraphics[width=0.85\textwidth]{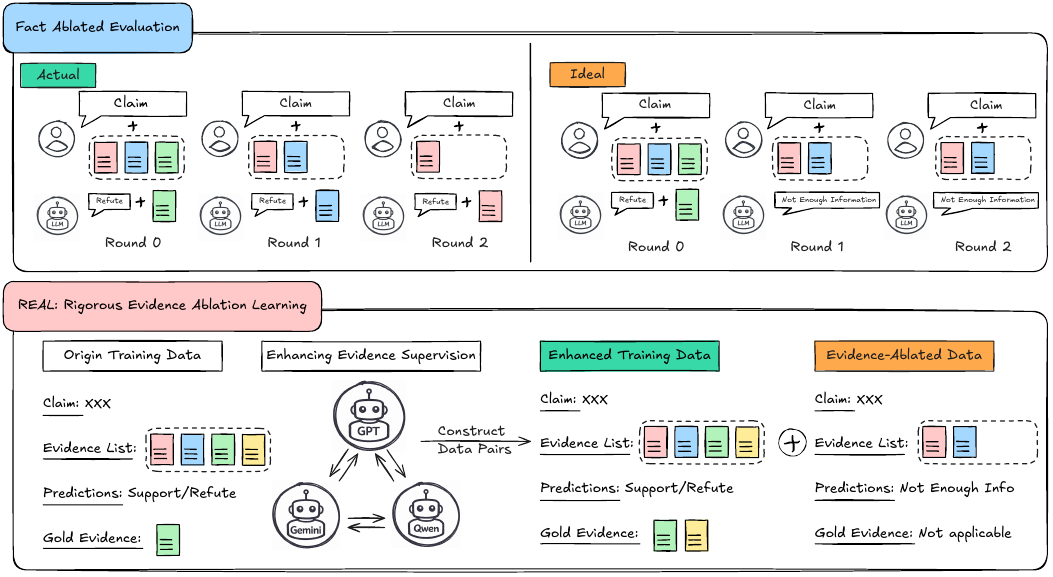}
    \caption{Illustration of proposed evaluating framework FAE (Top) and training framework REAL (Bottom)}
    \Description{The figure about the proposed FAE and REAL frameworks}
    \label{fig:FAE-REAL}
\end{figure*}

\section{Related Work}

The tendency of LLMs to memorise data observed during training is now well-established \cite{kiyomaru-etal-2024-comprehensive-analysis, wei2025memorization, di2025llms,carlini2021extracting}. 
Consequently, the adoption of RAG for such applications has necessitated extensive research into LLM grounding and refusal mechanisms~\cite{gao2023retrieval}.
This work can be broadly categorised into three directions. First, regarding robustness to irrelevant or insufficient context, several studies investigate how LLMs manage situations where retrieved documents are non-relevant or incomplete~\cite{yang2025knowing, thakur-etal-2024-knowing, joren2025sufficient, niu-etal-2024-ragtruth}. They emphasise the model's ability to ``know when it doesn't know'' and abstain from answering to prevent hallucinations. Second, in terms of evaluation frameworks, new metrics and benchmarks have been proposed to diagnose retrieval and generation errors~\cite{ru2024ragchecker, song2025measuring}, with a focus on measuring semantic consistency between answers and citations. Third, for methodological improvements, various adaptation and alignment strategies are developed to enhance grounding~\cite{ye-etal-2024-effective, hsu-etal-2024-calm, huang-etal-2024-learning}. For instance, Song et al.~\cite{song2025measuring} focus on aligning LLMs to abstain from answering under insufficient evidence, whereas Huang et al.~\cite{huang-etal-2024-learning} enhance verifiability by training models to ground claims in specific fine-grained textual evidence rather than coarse document identifiers.

While these works share our motivation to reduce reliance on internal parametric knowledge, they primarily operate in generation-centric settings, such as question-answering. In such contexts, grounding is typically defined as the semantic alignment between a free-form response and its cited sources. Consequently, while some evaluation frameworks~\cite{tang-etal-2024-minicheck, honovich-etal-2022-true-evaluating} utilise fact-checking related principles, they still tend to simplify the task by treating \texttt{REFUTE} and \texttt{NOT\_ENOUGH\_INFO} into a single \texttt{NON-SUPPORT} category.

In contrast, our work focuses on the task of fact-checking itself, which adopts a stricter formulation than generation-centric grounding by requiring granular labels and explicit causal dependency on evidence. Our work also differs from Akhtar et al.~\cite{akhtar2024ev2r} who removed gold evidence to validate an evaluation metric, but this static approach cannot diagnose whether a model actually uses that evidence for its decision. Instead, we introduce a dynamic ablation process that ablates the model's own predicted evidence. This allows us to characterise the causal dependency between the evidence a model selects and its final verdict, revealing when a model relies on parametric knowledge instead.

\section{Background}

\subsection{Problem Statement}
Given a set of sentences ($S$), a fact-checking system ($FC$)  assess a claim ($c$) and outputs a veracity label  ($l \in \mathcal{L}$) together with the supporting evidence sentences ($E \subseteq S$):
\begin{equation}
\label{eq:pipeline_equation_revised}
  FC(c, S) \rightarrow (l, E).
\end{equation}

$FC$ performs verification by maximising the joint probability of the target veracity label and the evidence subset:

\begin{equation}
\label{eq:joint_verifier_equation_pool}
\begin{aligned}
  (l, E)
  &= \underset{l \in \mathcal{L},\, E \subseteq S}{\arg\max}\;
     P(l, E \mid c, S)
\end{aligned}
\end{equation}

$\mathcal{L} = \{ {\tt SUPPORT}, {\tt REFUTE}, {\tt NOT\_ENOUGH\_INFO  / NEI}\}$ is a commonly used a set veracity labels \cite{guo-etal-2022-survey,zeng2021automated,vladika-matthes-2023-scientific}.

\subsection{Fact-Checking System Evaluation}
\label{ssec:eval_standard_fc}
Existing evaluation protocols for fact-checking systems primarily assess two independent aspects: (1) whether a model predicts the correct veracity label supported by identified evidence, and (2) whether the predicted evidence set covers the gold evidence.

\noindent(1) Veracity prediction is evaluated at the claim level, typically via two metrics \cite{guo-etal-2022-survey,zeng2021automated,wadden-etal-2020-fact,wang-etal-2023-check-covid,thorne-etal-2018-fever,schlichtkrull2023averitec}: (i) \textbf{Label Accuracy}, which measures the correctness of the predicted label $l$ against the gold label $l^*$. (ii) \textbf{Label-Evidence Joint (Strict) Accuracy}, which only credits a prediction if the veracity label is correct \textit{and} at least one gold evidence sentence is correctly identified.

\noindent(2) Evidence selection is usually evaluated by comparing the predicted evidence set $E$ against the annotated gold evidence $E^*$ from a \textbf{global perspective} \cite{guo-etal-2022-survey,zeng2021automated,wadden-etal-2020-fact,wang-etal-2023-check-covid,thorne-etal-2018-fever}. Three standard retrieval metrics are reported: \textbf{Precision}, \textbf{Recall}, and \textbf{$\mathbf{F_1}$ score}. These metrics are calculated by measuring the overlap between the set of all predicted evidence sentences and the set of all gold evidence.

\section{FAE: Fact Ablated Evaluation}
\subsection{Evidence Ablation Process}
\label{ssec:ablation_process}
To verify whether the model's predictions are strictly grounded in the provided evidence, we introduce Fact Ablated Evaluation (FAE), which evaluates model behaviour under varying levels of evidence availability. Figure~\ref{fig:FAE-REAL} (top) presents an overview of the FAE evaluation framework with an illustrative comparison between actual observations and the ideal performance of a fact-checking system. Given a claim $c$ and an evidence set $S$, $FC$ produces a prediction $(l^{(0)}, E^{(0)})$, referred to as the round $0$ predication. The evidence set is then ablated by removing the predicted evidence $E^{(0)}$:
\[
    S^{1} = S \setminus E^{(0)}.
\]
$FC$ then predicts the same claim $c$ again using $S^{1}$ to produce the round 1 prediction and then FAE updates the evidence set for round $2$:
\[
    (l^{(1)}, E^{(1)}),\ \ S^{2} = S^{1} \setminus E^{(1)}.
\]

This ablation process is repeated iteratively, dropping the predicted evidence $E^{(r)}$ following each round $r$ and re-evaluating $FC$ with evidence set $S^{r} \setminus E^{(r)}$ in the next round. The ablation process is terminated after a fixed number of rounds $R$. The resulting sequence $l^{(0)},\, l^{(1)},\, l^{(2)},\, \ldots \ l^{(r)}$ captures how $FC$ behaviour changes as cited evidence is progressively eliminated.

\subsection{Ideal vs. Actual Trajectories}
\label{sec:ideal}
\begin{figure}[h]
    \centering
    \includegraphics[width=0.60\linewidth]{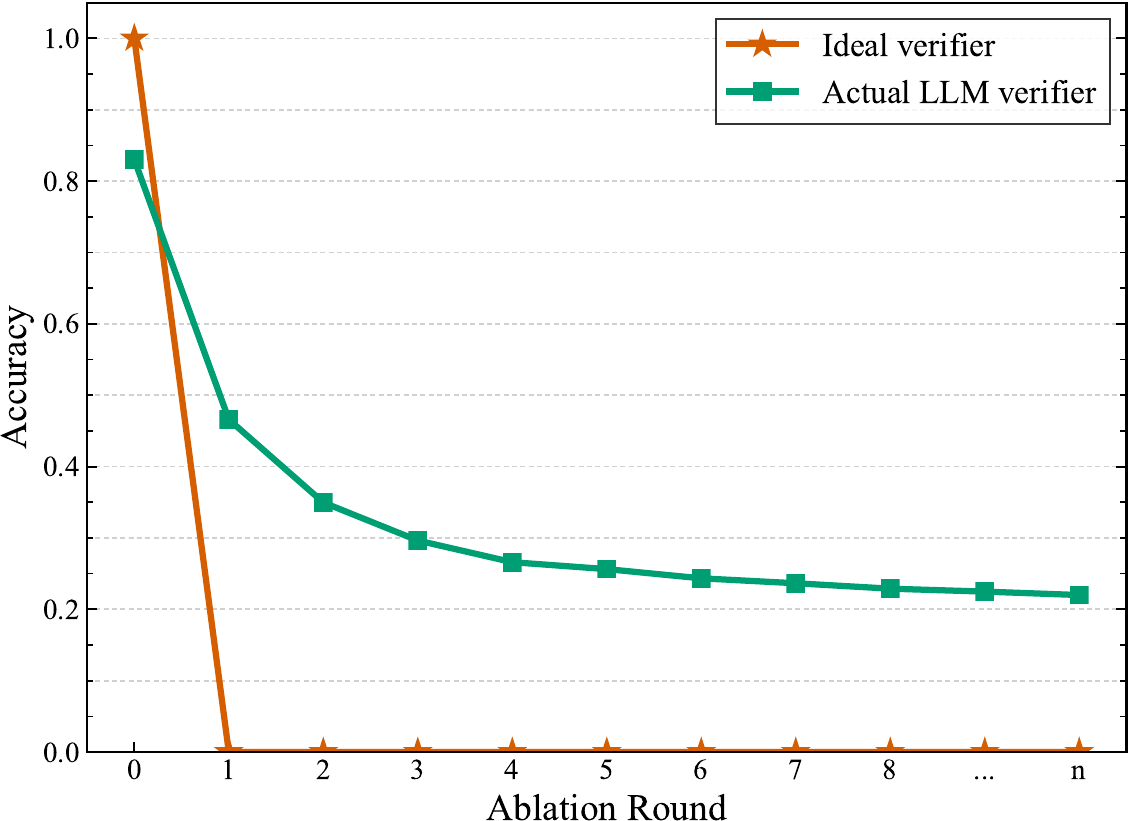}
    \caption{Ideal vs.\ actual ablation trajectories for claims annotated as \texttt{SUPPORT}/\texttt{REFUTE} (\texttt{Llama-3.1-8B-Instruct} on FEVER).}
    \label{fig:trajectory}
    \Description{The figure about Ideal vs. Actual Observation in FAE evaluation framework}
\end{figure}

\noindent After an initial \texttt{SUPPORT} or \texttt{REFUTE} prediction at round $r=0$, an ideal $FC$ should switch to \texttt{NOT\_ENOUGH\_INFO} once the supporting evidence has been ablated. However, experimental observations illustrated in Figure~\ref{fig:FAE-REAL} demonstrate that current LLM-based verifiers fail to exhibit this ideal behaviour, often maintaining non-\texttt{NEI} predictions across subsequent ablation rounds despite the unavailability of supporting evidence. Figure~\ref{fig:trajectory} further contrasts these ideal and observed trajectories over a sequence of ablation rounds. This observation suggests that the decisions of current LLM-based $FC$ do not rely solely on the provided evidence. Hence, we propose the following hypothesis: 

$\mathcal{H}_a$: \textit{Parametric knowledge acquired during training in LLMs interferes with the fidelity of strictly evidence-grounded verification.}

To test this hypothesis, we define two fact-checking paradigms: evidence-based and knowledge-based. Evidence-based paradigm aims to verify claims using only the provided evidence, while the knowledge-based paradigm aims to omit the evidence and prompt the LLM to rely completely on its parametric knowledge.

\begin{figure}[t]
    \centering
    \includegraphics[width=0.9\linewidth]{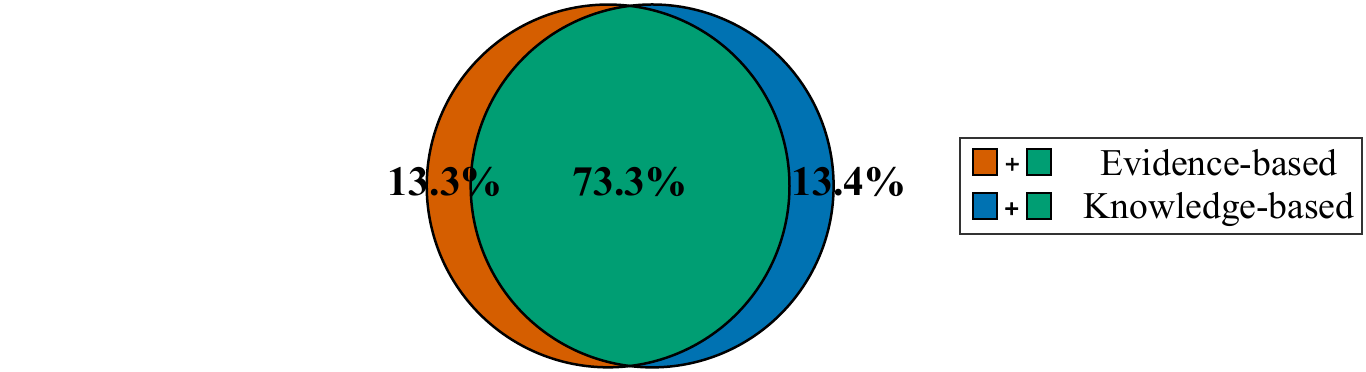}
    \caption{Comparison of activated neuron sets in fact checking using internal knowledge and provided evidence.}
    \Description{The Venn graph about activated neuron sets in fact checking using internal knowledge and provided evidence.}
    \label{fig:Venn}
    \vspace{-7pt}
\end{figure}

Following prior work \citep{tang-etal-2025-enhancing}, we record neuron activations on a calibration set. We rank each neuron by activation magnitude and select those that cumulatively account for $90\%$ of the total signal as highly associated neurons. This procedure is applied separately to the two paradigms to obtain paradigm-specific neuron sets, which are visualised in Figure~\ref{fig:Venn}. We observe that there is substantial overlap between the two paradigms ($73.3\%$), indicating shared fact-checking behaviour. However, the two paradigms also exhibit distinct paradigm-specific neuron subsets. This suggests that evidence-based and knowledge-based fact-checking are not identical at the neuronal level. Since the knowledge-based paradigm removes the evidence list, its specific neurons reflect the use of knowledge acquired during training. This supports $\mathcal{H}_a$ by showing that LLM verifiers retain a neuron-level signal associated with parametric knowledge, which may compete with the provided evidence when strictly evidence-grounded verification is required.

\subsection{FAE Metrics}\label{sec:fae_metrics}

To systematically evaluate behaviour under evidence ablation, we first compute accuracy at each ablation round, normalised by the accuracy obtained before any evidence ablation.
\begin{equation}
\label{eq:fae_gt}
g_r = \frac{a_r}{a_0} 
\end{equation}
where $a_r$ denotes the $FC$'s accuracy at ablation round $r$. The resulting sequence $\{g\} = \{g_0, g_1, \ldots, g_R\}$ describes how verification accuracy is retained as supporting evidence is progressively ablated.

Based on $\{g\}$, we further propose three evidence-grounded metrics. $IS$, $ER$, and $IO$ respectively capture the immediate response to evidence ablation, the persistence of non-\texttt{NEI} predictions at the final ablation round, and the cumulative deviation from ideal evidence-grounded behaviour.

\noindent{\textbf{Immediate Sensitivity ($IS$)}} measures the immediate response of a verifier’s predictions to evidence ablation:
\begin{equation}
\label{eq:fae_is}
\IS = 1 - g_1
\end{equation}

A high $\IS$ indicates an immediate collapse after the first ablation step, suggesting that the initial prediction depends directly on the cited evidence. Conversely, a low $\IS$ indicates that the prediction remains unchanged after initial evidence ablation, reflecting either incorrect evidence selection or limited reliance on the cited evidence.

\noindent{\textbf{End-State Retention ($ER$)}} measures behaviour at the final round:
\begin{equation}
\label{eq:fae_er}
ER = 1 - g_R
\end{equation}

A low $ER$ indicates persistent non-\texttt{NEI} predictions despite complete evidence ablation, a behaviour often associated with spurious or missing evidence, suggesting that the decision is driven by parametric knowledge rather than evidence dependence.

\noindent{\textbf{Ideal Offset ($IO$)}} measures the cumulative deviation from ideal evidence-grounded behaviour across the ablation process. Unlike $\IS$ and $ER$, which capture only the initial and final ablation rounds, $IO$ reflects behaviour dynamics across all ablation rounds.

\begin{equation}
\label{eq:fae_io}
IO
= 1 - \frac{1}{R} \sum_{r=1}^{R} a_r .
\end{equation}

This metric penalises $FC$ that remain confident as evidence availability degrades across intermediate ablation rounds and rewards a rapid, sustained performance drop following evidence ablation. We compute $IO$ using $a_r$ instead of $g_r$, without relying on $a_0$.

\section{REAL}
Using FAE, we observe that current LLM-based verifiers often maintain their original predictions even after the supporting evidence has been ablated, as illustrated in Figure~\ref{fig:trajectory}. This behaviour suggests that verification decisions are not fully grounded in the retrieved evidence, but are partially sustained through parametric knowledge acquired during pre-training. Such behaviour undermines evidence-dependent verification, since the final prediction can remain stable despite changes in the supporting context.

To address this limitation, we propose Rigorous Evidence Ablation Learning (\textbf{REAL}), a training framework designed to strengthen the dependency between verification decisions and evidence availability, shown in Figure~\ref{fig:FAE-REAL}. Unlike standard supervised fine-tuning (SFT), which optimises prediction correctness only under evidence-complete conditions, REAL additionally supervises the verifier's behaviour under evidence-ablated conditions where the supporting evidence has been intentionally removed. The key intuition is that a verifier should not only learn what prediction to make when sufficient evidence is available, but also when not to maintain the same prediction once the supporting evidence becomes unavailable.

\subsection{Evidence Ablation}
\label{sec:Evidence_Ablation}

Previous work on LLM-based fact-checking typically fine-tunes models using pairs consisting only of a claim and its supporting evidence \cite{shcharbakova-etal-2025-scale, cheung2023factllama, kumar2025improving, putta-etal-2025-claimcheck, yoon-etal-2024-hero}. Under this paradigm, the verifier is exposed exclusively to evidence-complete training conditions, where the required evidence is always available together with the correct label. Consequently, the model is never explicitly trained to perform evidence-grounded verification when supporting evidence becomes unavailable.

As a result, the verifier is not trained to change its prediction when supporting evidence is removed. During training, maintaining the original prediction after evidence ablation is never penalised as long as the prediction remains correct under the original evidence-complete input. Consequently, the verifier may learn to preserve predictions through parametric knowledge instead of adapting its behaviour according to evidence availability.

To address this issue, REAL represents each training instance as a triplet $(c, S, l)$, where $c$ is a claim, $S$ is the sentence set, and $E \subseteq S$ denotes the gold evidence supporting the gold label $l^*$. For each training instance, REAL constructs two contrasting supervision conditions:

\begin{itemize}[leftmargin=7pt]
\item \textbf{Positive Input:} Full context $S$, with target output $(l, E)$.
\item \textbf{Negative Input:} Ablated context $S_{\text{abl}} = S \setminus E^*$, with target output $(\texttt{NEI}, \emptyset)$, as shown in the bottom-right of Figure~\ref{fig:FAE-REAL}.
\end{itemize}

For example, as illustrated in Figure~\ref{fig:failure_example}, if the claim ``Queen is a British rock band formed in London'' is supported by 'Sentence [1]', the ablated input removes this supporting sentence while preserving the remaining document context. REAL then supervises the verifier to abstain rather than maintain the original \texttt{SUPPORT} prediction through unrelated or hallucinated evidence.

By jointly supervising these two contrasting conditions, REAL converts evidence availability into a counterfactual supervision signal. The verifier is therefore required not only to produce correct predictions under sufficient evidence, but also to revise its behaviour once the supporting evidence becomes unavailable.

Formally, REAL optimises the following objective:
\vspace{-7pt}

\begin{equation}
\mathcal{L}_{REAL}
=
-
\log P_{\theta}(l^*, E^* \mid c, S)
-
\log P_{\theta}(\texttt{NEI}, \emptyset \mid c, S_{\text{abl}})
\end{equation}

\noindent where the first term supervises standard evidence-grounded verification under complete evidence conditions, and the second term penalises prediction persistence after the supporting evidence is removed. This paired objective encourages the verifier to produce a veracity prediction only when sufficient evidence is available, and to abstain when the evidence required for that prediction is no longer present.

\subsection{Evidence Enhancement}
\label{sec:Evidence_Enhance}

The construction of evidence-ablated inputs assumes that the gold evidence set exhaustively captures all sentences in $S$ that support or refute the claim. However, evidence annotations in fact-checking benchmarks, including FEVER \cite{thorne-etal-2018-fever}, are typically designed for sufficiency rather than exhaustiveness \cite{derczynski-etal-2020-maintaining, bekoulis2021review}. In practice, multiple valid evidence sentences may coexist within the same document even though only a subset is annotated.

This incompleteness introduces a critical challenge for evidence-ablated supervision. Positive samples may omit valid supporting evidence, while negative samples constructed through evidence ablation may still contain unannotated evidence capable of supporting the original claim. Consequently, logically valid verifier predictions may be incorrectly penalised during training, weakening the reliability of the counterfactual supervision signal.

To mitigate this issue, REAL augments the original evidence annotations through cross-model evidence verification. Specifically, we use GPT-4o-mini \cite{hurst2024gpt}, Gemini-2.5-Flash-Lite \cite{comanici2025gemini}, and Qwen-2.5-32B-Instruct \cite{qwen2025qwen25technicalreport} to independently identify supporting evidence sentences from the sentence pool $S$ for each claim. Each model independently identifies supporting evidence, and we retain sentences supported by the majority of models to construct the augmented evidence set. By aggregating evidence predictions across multiple LLMs, this process reduces the likelihood of leaving valid supporting evidence inside ablated contexts. This helps mitigate incorrect counterfactual supervision caused by incomplete evidence annotations during REAL training.

\section{Experiments}
\subsection{Models}

\noindent\textbf{\textit{Baselines.}} 
A wide range of LLMs have been adopted as verifiers for verdict prediction in fact-checking (e.g. GPT$\ast$ \cite{rothermel-etal-2024-infact,braun2025defame,ullrich-etal-2024-aic,park-etal-2024-dunamu,jannah-etal-2025-multilingual,li2025use,wang-etal-2024-factcheck,xie-etal-2025-fire,wei2024longform,geng2025m4fc}, Llama$\ast$ \cite{yoon-etal-2024-hero, mohammadkhani-etal-2024-zero,zhou-etal-2025-gqc,xie-etal-2025-fire,geng2025m4fc}, Qwen$\ast$ \cite{yoon-etal-2025-team,ullrich-drchal-2025-aic,zhou-etal-2025-gqc,geng2025m4fc} and Gemini$\ast$ \cite{jannah-etal-2025-multilingual,chowdhury-etal-2025-fact5,wei2024longform,geng2025m4fc}), leveraging their capability to reason over claims and supporting evidence for veracity prediction. However, these studies mainly employ the models in a \textbf{zero-shot} or \textbf{few-shot} manner through prompting, without task-specific fine-tuning to ensure evidence-grounded consistency. Hence, we evaluate REAL against two categories of baselines to demonstrate its effectiveness in veracity prediction and grounding consistency:

\textbf{Off-the-shelf LLMs} (\S\ref{ssec:result_fever}, \S\ref{ssec:result_model_ablation}): We include proprietary LLMs (GPT-4o-mini \cite{hurst2024gpt}, Gemini-2.5-Flash-lite \cite{comanici2025gemini}) and open-source models (Qwen-2.5-32B/7B-Instruct \cite{qwen2025qwen25technicalreport}, Llama-3.1-8B-Instruct\cite{grattafiori2024llama}) to evaluate fact-checking behaviour without task-specific fine-tuning (\S\ref{ssec:result_fever}, \S\ref{ssec:result_more_datasets}  ).

\textbf{Fine-tuned LLMs} (\S\ref{ssec:result_training_ablation}): We also compare against the standard supervised fine-tuning following a few prior studies which applied this approach to fact-checking \cite{shcharbakova-etal-2025-scale, cheung2023factllama, kumar2025improving, putta-etal-2025-claimcheck, yoon-etal-2024-hero} by directly leveraging supervision data provided by claim and gold evidence pairs only (Table~\ref{tab:ablation_real}, \S\ref{ssec:result_model_ablation} ).

\noindent\textbf{\textit{Base LLMs for REAL.}} 
Following prior work~\cite{shcharbakova-etal-2025-scale,cheung2023factllama,kumar2025improving,putta-etal-2025-claimcheck,yoon-etal-2024-hero}, we adopt Llama-3.1-8B-Instruct \cite{grattafiori2024llama} as the primary backbone of REAL. To demonstrate framework generality, we also evaluate REAL under Qwen-2.5-7B-Instruct \cite{qwen2025qwen25technicalreport} in an ablation study (Table~\ref{tab:ablation_model_family}, \S\ref{ssec:result_model_ablation} ).

\subsection{Hyperparameter Details}
Models are fine-tuned using LoRA~\cite{hu2022lora} with rank $r=16$, scaling factor $\alpha=32$, and dropout $0.05$, applied to the attention projection. Training is performed using AdamW \cite{loshchilov2018decoupled} with a learning rate of $1\times10^{-4}$, and a global batch size of 32. The maximum input sequence length is set to 4096, which covers total length of all pairs for the used datasets. Models are fine-tuned using a supervised assistant-only objective~\cite{ouyang2022training}, to generate structured outputs consisting of a veracity label and evidence sentence indices. Models are trained for two epochs, with a fixed random seed across all experiments. Code for data preparation and experiments is available in github\footnote{\url{https://github.com/xingyu-deng/FAE_REAL}}.

\subsection{Datasets}
This work includes four datasets for in-domain (FEVER) and out-of-domain evaluation (SciFact, Climate-FEVER and Check-COVID):
\begin{itemize}[leftmargin=7pt] 
\item \textbf{FEVER} \cite{thorne-etal-2018-fever} consists of 185,445 human-written claims verified against Wikipedia articles, with sentence-level annotations.
\item \textbf{SciFact} \cite{wadden-etal-2020-fact} contains 1,409 scientific claims verified against research paper abstracts, with domain expert evidence annotations.
\item \textbf{Climate-FEVER} \cite{diggelmann2020climate} is a domain-specific benchmark focused on climate-related claims, verified against Wikipedia.
\item \textbf{Check-COVID} \cite{wang-etal-2023-check-covid} consists of COVID-19 related claims verified against scientific and medical sources, representing a challenging real-world verification setting.
\end{itemize}

Claims labelled as \texttt{NEI} require an external retrieval process to obtain documents, which introduces unnecessary variables (e.g., retriever quality and label noise) that are outside the scope of this study, and are therefore excluded due to their limited relevance to the core research focus of this work.

\textbf{\textit{Training.}} We train LLMs under proposed REAL framework on the train split of FEVER \cite{thorne-etal-2018-fever} only, as it provides sufficient scale for generalisable evidence-grounding behaviour. 

\textbf{\textit{Evaluation.}} 
We evaluate LLM-based verifiers under two settings. For \textbf{in-domain evaluation}, we test on the FEVER test split. To assess generalisation ability, we conduct \textbf{out-of-domain evaluation} on SciFact \cite{wadden-etal-2020-fact}, Climate-FEVER \cite{diggelmann2020climate}, and Check-COVID \cite{wang-etal-2023-check-covid} without additional training, using all available splits. For SciFact, we exclude the test split, as it is blind for the shared task \cite{wadden-lo-2021-overview}.

\subsection{Evaluation Metrics}
\textbf{\textit{Standard fact-checking metrics.}}
Following standard protocols, as defined in \S\ref{ssec:eval_standard_fc}, we evaluate the initial prediction using \textbf{Label Accuracy} and \textbf{Strict Accuracy} for veracity prediction, and \textbf{Precision}, \textbf{Recall}, and \textbf{$F_1$ Score} for evidence selection.

\noindent\textbf{\textit{Fact Ablated Evaluation metrics.}}
We assess evidence dependence using the FAE metrics (\textbf{IS}, \textbf{ER} and \textbf{IO}) described in \S\ref{sec:fae_metrics} over four ablation rounds following the initial prediction ($R=4$ in Equation~\ref{eq:fae_gt}). We set $R=4$ based on our empirical observations that the verifier's accuracy converges to a stable plateau by this stage. Beyond this point, continued ablation of the residual context adds minimal information regarding the model's evidence dependence. Instead, the verifier often fails by hallucinating non-existing evidence identifiers (the third failure mode in Figure~\ref{fig:failure_example}).

\begin{figure*}[t]
    \centering
    \includegraphics[width=0.93\linewidth]{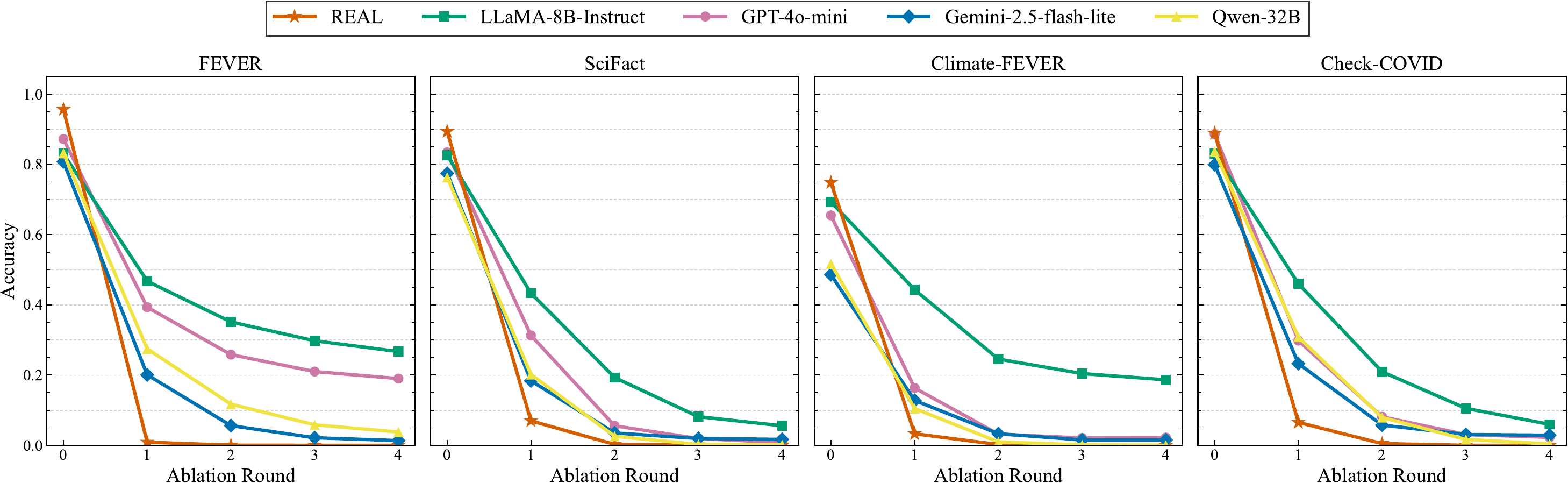}
    \caption{Ablation trajectories in FEVER, SciFact, Climate-FEVER and Check-COVID. }
    \Description{The figure about the experiment results in FEVER, SciFact, Climate-FEVER and Check-COVID.}
    \label{fig:fae_graph}
\end{figure*}

\begin{table*}[h]
\centering
\caption{Fact-Checking performance (left) and FAE metrics (right) in four datasets}
\begin{tabular}{llccclccccc}
\hline
\multirow{2}{*}{\textbf{Model/Metrics}} &  & \multicolumn{6}{c}{\textbf{Fact-Checking Metrics (Initial)}}                                  & \multicolumn{3}{c}{\multirow{2}{*}{\textbf{FAE Metrics}}} \\ \cline{3-5} \cline{7-8}
                                        &  & \multicolumn{3}{c}{Evidence Selection}           &  & \multicolumn{2}{c}{Veracity prediction} & \multicolumn{3}{c}{}                                      \\ \hline
\textbf{FEVER}                          &  & Prec           & Rec            & F1             &  & Acc                & Strict             & $IS$              & $ER$              & $IO$              \\ \hline
GPT-4o-mini                             &  & 74.46          & {\ul 70.09}    & {\ul 72.21}    &  & {\ul 87.24}        & {\ul 81.97}        & 54.91             & 78.21             & 71.27             \\
Gemini-2.5-flash-lite                   &  & {\ul 75.69}    & 66.66          & 70.89          &  & 80.78              & 78.46              & {\ul 75.15}       & {\ul 98.33}       & {\ul 90.72}       \\
Qwen-2.5-32B-Instruct                   &  & \textbf{79.51} & 65.17          & 71.63          &  & 83.19              & 78.81              & 66.95             & 95.45             & 84.98             \\
Llama-3.1-8B-Instruct                   &  & 56.30          & 69.40          & 62.17          &  & 83.08              & 76.27              & 43.79             & 67.85             & 62.83             \\
REAL (ours)                                 &  & 71.64          & \textbf{85.89} & \textbf{78.12} &  & \textbf{95.66}     & \textbf{94.28}     & \textbf{99.06}    & \textbf{99.99}    & \textbf{99.69}    \\ \hline
\textbf{SciFact}                        &  & Prec           & Rec            & F1             &  & Acc                & Strict             & $IS$              & $ER$              & $IO$              \\ \hline
GPT-4o-mini                             &  & 66.41          & \textbf{74.62} & {\ul 70.28}    &  & {\ul 83.44}        & {\ul 82.54}        & 62.84             & 99.07             & 87.06             \\
Gemini-2.5-flash-lite                   &  & 64.72          & 63.05          & 63.87          &  & 77.44              & 74.58              & {\ul 76.26}       & 97.81             & 92.05             \\
Qwen-2.5-32B-Instruct                   &  & \textbf{69.76} & 62.69          & 66.04          &  & 76.33              & 73.22              & 73.56             & {\ul 99.83}       & {\ul 92.32}       \\
Llama-3.1-8B-Instruct                   &  & 63.75          & 67.51          & 65.57          &  & 82.66              & 77.88              & 47.72             & 93.25             & 76.48             \\
REAL (ours)                                 &  & {\ul 68.80}    & {\ul 72.97}    & \textbf{70.82} &  & \textbf{89.39}     & \textbf{84.86}     & \textbf{92.19}    & \textbf{100}      & \textbf{97.54}    \\ \hline
\textbf{Climate-FEVER}                  &  & Prec           & Rec            & F1             &  & Acc                & Strict             & $IS$              & $ER$              & $IO$              \\ \hline
GPT-4o-mini                             &  & {\ul 74.01}    & 57.78          & {\ul 64.90}    &  & 65.49              & 62.62              & 75.08             & 96.63             & 92.80             \\
Gemini-2.5-flash-lite                   &  & 76.98          & 38.15          & 51.02          &  & 48.62              & 46.20              & 73.70             & {\ul 96.83}       & 94.12             \\
Qwen-2.5-32B-Instruct                   &  & \textbf{78.32} & 42.97          & 55.50          &  & 51.71              & 49.28              & {\ul 79.74}       & \textbf{100}      & {\ul 96.14}       \\
Llama-3.1-8B-Instruct                   &  & 70.04          & {\ul 59.95}    & 64.60          &  & {\ul 69.35}        & {\ul 63.40}        & 37.14             & 72.94             & 70.90             \\
REAL (ours)                                 &  & 71.39          & \textbf{62.56} & \textbf{66.68} &  & \textbf{74.86}     & \textbf{68.69}     & \textbf{95.58}    & \textbf{100}      & \textbf{98.82}    \\ \hline
\textbf{Check-COVID}                  &  & Prec           & Rec            & F1             &  & Acc                & Strict             & $IS$              & $ER$              & $IO$              \\ \hline
GPT-4o-mini                             &  & \textbf{54.12} & \textbf{74.85} & \textbf{62.82} &  & {\ul 88.60}        & \textbf{87.02}     & 66.33             & 97.43             & 86.32             \\
Gemini-2.5-flash-lite                   &  & 54.10          & 62.06          & 57.80          &  & 79.94              & 73.49              & {\ul 70.86}       & 96.39             & {\ul 89.20}       \\
Qwen-2.5-32B-Instruct                   &  & 53.20          & 63.70          & 57.98          &  & 83.71              & 76.53              & 68.89             & {\ul 99.74}       & 89.17             \\
Llama-3.1-8B-Instruct                   &  & 53.26          & 63.58          & 57.96          &  & 83.05              & 72.35              & 44.72             & 92.81             & 74.30             \\
REAL (ours)                                 &  & {\ul 54.10}    & {\ul 72.50}    & {\ul 61.97}    &  & \textbf{88.90}     & {\ul 83.65}        & \textbf{92.64}    & \textbf{100}      & \textbf{97.65}    \\ \hline
\end{tabular}
\label{tab:results_all}
\end{table*}

\subsection{Research Questions}

To evaluate the effectiveness of REAL and analyse its underlying mechanisms, we address the following research questions:
\begin{itemize}[leftmargin=7pt]

\item \textbf{RQ1: Veracity and Grounding.} Does REAL improve veracity accuracy while enforcing strict evidence grounding? 
We address this by comparing REAL against baselines on the in-domain evaluation benchmark (i.e., FEVER)
(\S\ref{ssec:result_fever}).

\item \textbf{RQ2: Domain Generalisation.} Does the evidence-dependent behaviour learned by REAL transfer to out-of-domain settings? 
We evaluate this on the out-of-domain benchmarks 
(\S\ref{ssec:result_more_datasets}).

\item \textbf{RQ3: Ablation Studies.} How do individual components in REAL contribute to accuracy, and does REAL generalise across model architectures? 
We answer this via detailed ablation studies 
(\S\ref{ssec:framework_generality}).

\item \textbf{RQ4: Internal Mechanism.} What internal representational changes underpin evidence-dependent behaviour in LLMs fine-tuned by REAL? 
We analyse this by extracting the dynamics of representation space during evidence ablation 
(\S\ref{ssec:result_representation}).
\end{itemize}

\section{Results}

\subsection{Veracity and Grounding (RQ1)}
\label{ssec:result_fever}

\begin{table*}[h]
\centering
\caption{Ablation study for proposed REAL training framework.}
\begin{tabular}{llccclccccc}
\hline
\multirow{2}{*}{\textbf{Model/Metrics}} &  & \multicolumn{6}{c}{\textbf{Fact-Checking Metrics (Initial)}}                           & \multicolumn{3}{c}{\multirow{2}{*}{\textbf{FAE Metrics}}} \\ \cline{3-5} \cline{7-8}
                                        &  & \multicolumn{3}{c}{Evidence Selection} &     & \multicolumn{2}{c}{Veracity prediction} & \multicolumn{3}{c}{}                                                 \\ \hline
\textbf{FEVER}                          &  & Prec         & Rec        & F1         &     & Acc                & Strict             & $IS$                  & $ER$                  & $IO$                 \\ \hline
Standard SFT              &  & 75.45        & 85.10      & 79.99      &     & 96.18              & 93.87              & 10.75                 & 11.02                 & 14.36                \\
w/o Evidence Ablation (\S~\ref{sec:Evidence_Ablation})                           &  & 71.14        & 86.61      & 78.12      &     & 96.06              & 94.52              & 10.80                 & 11.04                 & 14.67                \\
w/o Evidence Enhancement (\S~\ref{sec:Evidence_Enhance})                        &  & 77.90        & 83.29      & 80.50      &     & 94.28              & 92.24              & 98.82                 & 99.99                 & 99.61                \\
REAL                                 &  & 71.64        & 85.89      & 78.12      &     & 95.66              & 94.28              & 99.06                 & 99.99                 & 99.69                \\ \hline
\textbf{SciFact}                        &  & Prec         & Rec        & F1         &     & Acc                & Strict             & $IS$                  & $ER$                  & $IO$                 \\ \hline
Standard SFT              &  & 71.95        & 42.77      & 53.65      &     & 90.04              & 68.56              & 9.77                  & 15.37                 & 22.81                \\
w/o Evidence Ablation (\S~\ref{sec:Evidence_Ablation})                           &  & 64.87        & 75.27      & 69.68      &     & 91.07              & 86.16              & 14.35                 & 21.45                 & 25.49                \\
w/o Evidence Enhancement (\S~\ref{sec:Evidence_Enhance})                        &  & 70.65        & 42.06      & 52.73      &     & 87.97              & 67.40              & 92.79                 & 100                   & 97.46                \\
REAL                                 &  & 68.80        & 72.97      & 70.82      &     & 89.39              & 84.86              & 92.19                 & 100                   & 97.54                \\ \hline
\textbf{Climate-FEVER}                  &  & Prec         & Rec        & F1         &     & Acc                & Strict             & $IS$                  & $ER$                  & $IO$                 \\ \hline
Standard SFT              &  & 69.86        & 32.89      & 44.72      &     & 84.45              & 61.85              & 3.92                  & 8.09                  & 20.03                \\
w/o Evidence Ablation (\S~\ref{sec:Evidence_Ablation})                           &  & 69.88        & 63.70      & 66.65      &     & 84.79              & 74.86              & 7.68                  & 7.68                  & 22.47                \\
w/o Evidence Enhancement (\S~\ref{sec:Evidence_Enhance})                        &  & 68.69        & 29.49      & 41.26      &     & 77.40              & 56.56              & 93.30                 & 100                   & 97.98                \\
REAL                                 &  & 71.39        & 62.56      & 66.68      &     & 74.86              & 68.69              & 95.58                 & 100                   & 98.82                \\ \hline 
\textbf{Check-COVID}                    &  & Prec         & Rec        & F1         &     & Acc                & Strict             & $IS$                  & $ER$                  & $IO$                 \\ \hline
Standard SFT              &  & 67.84        & 50.23      & 57.73      &     & 87.91              & 66.70              & 9.24                  & 18.26                 & 23.82                \\
w/o Evidence Ablation (\S~\ref{sec:Evidence_Ablation})                           &  & 55.94        & 73.24      & 63.43      &     & 91.08              & 85.53              & 13.10                 & 19.54                 & 24.09                \\
w/o Evidence Enhancement (\S~\ref{sec:Evidence_Enhance})                        &  & 68.83        & 48.42      & 56.85      &     & 83.45              & 62.83              & 93.82                 & 100                   & 98.12                \\
REAL                                 &  & 58.95        & 72.50      & 65.03      &     & 88.90              & 83.65              & 92.64                 & 100                   & 97.65                \\ \hline
\end{tabular}
\label{tab:ablation_real}
\end{table*}

Table~\ref{tab:results_all} reports standard fact-checking metrics and grounding-oriented FAE metrics on FEVER for in-domain evaluation. We observe that REAL consistently outperforms all other baselines across both verification accuracy and evidence-grounded behaviour. On veracity prediction, compared to Llama-3.1-8B-Instruct, REAL improves label accuracy from $83.08$ to $95.66$, and strict accuracy from $76.27$ to $94.28$. Interestingly, despite being built on Llama-3.1-8B-Instruct, REAL surpasses substantially larger proprietary and open-source LLMs across both fact-checking and FAE metrics. For example, compared with GPT-4o-mini, REAL improves strict accuracy from $81.97$ to $94.28$, while increasing the $IS$ score from $54.91$ to $99.06$. Compared with Gemini-2.5-Flash-Lite, REAL further improves the $IO$ score from $90.72$ to $99.69$. Regarding evidence selection, REAL achieves the best recall ($85.89$) and F1 ($78.12$) among baselines, indicating more complete and accurate identification of supporting evidence.

Importantly, the improvements in evidence dependency do not come at the expense of fact-checking performance. REAL simultaneously achieves stronger grounding behaviour and higher verification accuracy than all baselines. In contrast, several baseline models maintain relatively high verification performance even after substantial evidence ablation, indicating that strong initial accuracy alone may not fully reflect genuine evidence dependency during verification.

Regarding FAE metrics, we observe a clear difference in evidence-grounded behaviour. 
Baselines retain relatively high accuracy after evidence ablation, whereas REAL exhibits substantially stronger sensitivity to evidence availability. 
REAL achieves an $IS$ of $99.06$ and an $IO$ of $99.69$, compared to $75.15$ and $90.72$ for the best baseline (Gemini-2.5-Flash-Lite). 
Consistently, Figure~\ref{fig:fae_graph} shows that REAL’s accuracy rapidly collapses after the first ablation round, closely approaching the ideal evidence-grounded trajectory illustrated in Figure~\ref{fig:trajectory}. 
In contrast, baseline models stabilise at non-zero performance plateaus, indicating persistent prediction behaviour despite the removal of selected evidence.

In summary, the above results demonstrate that REAL improves both veracity accuracy and evidence-dependent behaviour for in-domain evaluation, addressing RQ1. By enforcing strict reliance on retrieved evidence, REAL achieves the best evidence-grounded scores, effectively mitigating the influence of parametric knowledge in standard LLMs for fact-checking while preserving strong prediction accuracy.

\begin{table*}[h]
\centering
\caption{Ablation study for proposed REAL trained with model from Qwen family}
\begin{tabular}{llccclccccc}
\hline
\multirow{2}{*}{\textbf{Model/Metrics}} &  & \multicolumn{6}{c}{\textbf{Fact-Checking Metrics (Initial)}}                           & \multicolumn{3}{c}{\multirow{2}{*}{\textbf{FAE Metrics}}} \\ \cline{3-5} \cline{7-8}
                                        &  & \multicolumn{3}{c}{Evidence Selection} &     & \multicolumn{2}{c}{Veracity prediction} & \multicolumn{3}{c}{}                                                 \\ \hline
\textbf{FEVER}                          &  & Prec         & Rec        & F1         &     & Acc                & Strict             & $IS$                  & $ER$                  & $IO$                 \\ \hline
Qwen-2.5-7B-Instruct                    &  & 70.82        & 67.40      & 69.07      &     & 81.94              & 77.18              & 59.95                 & 83.49                 & 77.11                \\
+REAL                                 &  & 81.40        & 79.73      & 80.56      &     & 95.23              & 93.43              & 99.04                 & 100                   & 99.66                \\ \hline
\textbf{SciFact}                        &  & Prec         & Rec        & F1         &     & Acc                & Strict             & $IS$                  & $ER$                  & $IO$                 \\ \hline
Qwen-2.5-7B-Instruct                    &  & 65.46        & 64.99      & 65.22      &     & 76.84              & 73.87              & 62.46                 & 97.14                 & 86.37                \\
+REAL                                 &  & 67.03        & 67.22      & 67.12      &     & 89.52              & 81.76              & 86.71                 & 99.86                 & 95.73                \\ \hline
\textbf{Climate-FEVER}                  &  & Prec         & Rec        & F1         &     & Acc                & Strict             & $IS$                  & $ER$                  & $IO$                 \\ \hline
Qwen-2.5-7B-Instruct                    &  & 70.25        & 53.45      & 60.71      &     & 63.29              & 58.43              & 55.40                 & 82.40                 & 82.40                \\
+REAL                                 &  & 71.64        & 51.85      & 60.13      &     & 78.83              & 66.70              & 94.97                 & 99.86                 & 98.60                \\ \hline
\textbf{Check-COVID}                    &  & Prec         & Rec        & F1         &     & Acc                & Strict             & $IS$                  & $ER$                  & $IO$                 \\ \hline
Qwen-2.5-7B-Instruct                    &  & 52.86        & 66.40      & 58.86      &     & 83.94              & 77.30              & 53.13                 & 93.15                 & 77.96                \\
+REAL                                 &  & 56.82        & 69.80      & 62.68      &     & 90.49              & 83.05              & 90.14                 & 99.89                 & 96.73                \\ \hline
\end{tabular}
\label{tab:ablation_model_family}
\end{table*}

\subsection{Domain Generalisation (RQ2)}
\label{ssec:result_more_datasets}

Table~\ref{tab:results_all} further reports out-of-domain evaluation results on SciFact, Climate-FEVER, and Check-COVID. REAL consistently maintains strong fact-checking performance while achieving the strongest FAE scores ($IS$, $ER$, and $IO$) across all evaluation domains, demonstrating robust transfer of evidence-grounded behaviour beyond the FEVER training distribution.

Importantly, the improvements in evidence dependency do not come at the expense of fact-checking performance. REAL achieves the best label accuracy and strict accuracy on SciFact ($89.39/84.86$) and Climate-FEVER ($74.86/68.69$), while remaining competitive on Check-COVID. More broadly, REAL is the only framework that consistently maintains both strong verification accuracy and strong evidence dependency across all evaluation datasets.

Among baselines, a clear trade-off emerges between evidence-grounded behaviour and fact-checking performance. Qwen-2.5-32B-Instruct and Gemini-2.5-Flash-Lite exhibit relatively stronger evidence dependency but weaker verification accuracy, whereas GPT-4o-mini and Llama-3.1-8B-Instruct achieve stronger initial prediction performance while remaining substantially less sensitive to evidence ablation. This observation suggests that stronger evidence-grounded behaviour does not naturally emerge from model scale or general reasoning capability alone, but instead requires explicit supervision over evidence availability.

The transfer behaviour is further visualised in Figure~\ref{fig:fae_graph}. Unlike the immediate collapse observed on FEVER, REAL’s accuracy on out-of-domain datasets approaches zero after approximately two ablation rounds. Although domain mismatch and fragmented evidence distributions prevent perfectly ideal collapse trajectories, REAL still maintains substantially stronger evidence sensitivity than all baselines throughout the ablation process.

Overall, these results demonstrate that the evidence-grounded behaviour learned through REAL generalises effectively across domains, addressing RQ2. REAL enforces a strict dependence on retrieved evidence, enabling a strong transfer of evidence-grounded behaviour, yielding a more reliable verifier in general, even in out-of-domain settings.

\subsection{Ablation Studies (RQ3)}
\label{ssec:framework_generality}

This section examines the effectiveness of REAL from two perspectives: the contribution of its individual components and its generalisation capability across model families.

\noindent \textbf{\textit{Different Design Choices.}}
\label{ssec:result_training_ablation}

\noindent
To assess the contribution of each component in REAL, we conduct ablation studies (Table~\ref{tab:ablation_real}) by isolating the effects of Evidence Ablation (\S\ref{sec:Evidence_Ablation}) and Evidence Enhancement (\S\ref{sec:Evidence_Enhance}).

\begin{figure*}[t]
\centering
  \begin{minipage}{0.95\textwidth}
    \includegraphics[width=\linewidth]{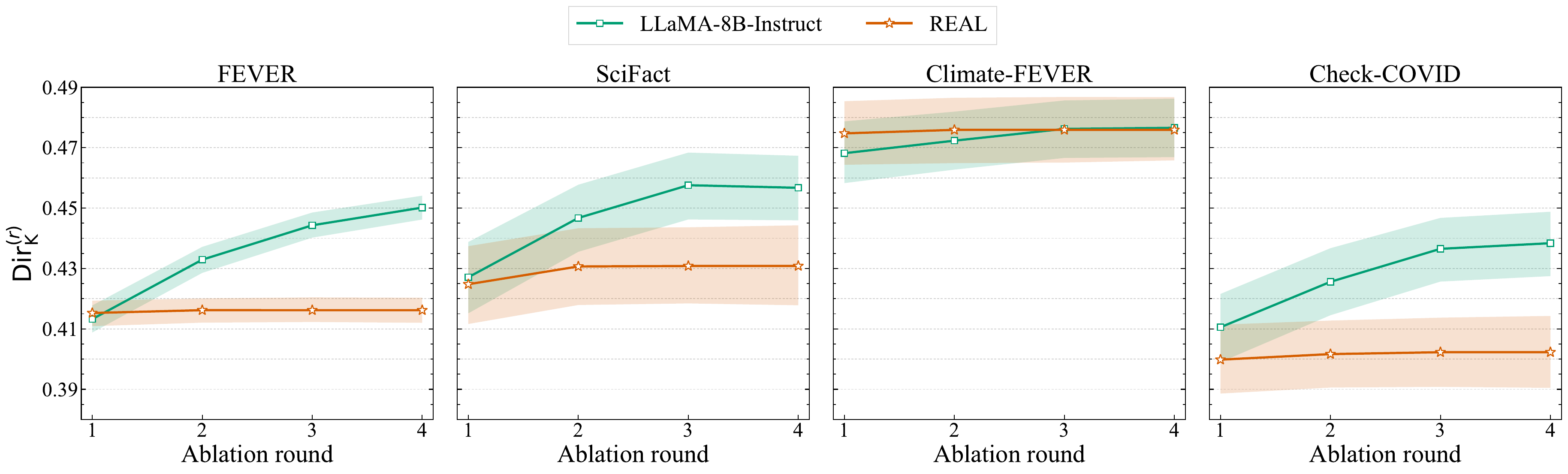}
    \caption{Directional alignment to knowledge-based fact-checking.}
    \Description{Directional alignment to knowledge-based fact-checking on FEVER, SciFact, Climate-FEVER and Check-COVID.}
    \label{fig:dire_to_zero}
  \end{minipage}
\end{figure*}

Evidence Ablation is the primary driver of evidence-grounded behaviour, revealing a clear disconnect between veracity prediction accuracy and evidence dependency. Removing this component causes FAE scores to collapse across all datasets despite relatively stable initial fact-checking performance. On FEVER, the $IS$ score drops from $99.06$ to $10.75$ under Standard SFT and to $10.80$ without Evidence Ablation. Similar degradation is consistently observed across all out-of-domain benchmarks. This observation aligns with the motivation of REAL: standard supervised fine-tuning alone does not constrain verifier behaviour under missing-evidence conditions, allowing prediction persistence through parametric knowledge.

In contrast, Evidence Enhancement primarily improves supervision quality rather than grounding behaviour itself. Removing this component while retaining Evidence Ablation preserves strong FAE scores, but substantially degrades fact-checking performance, particularly on SciFact and Climate-FEVER. For example, on Climate-FEVER, the evidence selection F1 score decreases from $66.68$ to $41.26$ without Evidence Enhancement. This degradation arises because incomplete evidence annotations leave residual supporting evidence inside ablated contexts, introducing incorrect counterfactual supervision signals during training.

Taken together, the two components of REAL play distinct and complementary roles. The Evidence Ablation improves evidence-grounded verification by constraining reliance on parametric knowledge when evidence is absent. The Evidence Enhancement, in contrast, improves veracity prediction and evidence selection under evidence-grounded behaviour.

\noindent\textbf{\textit{Different Model Families.}}
\label{ssec:result_model_ablation}
We further evaluate the generality of REAL by applying it to Qwen-2.5-7B-Instruct. As shown in Table~\ref{tab:ablation_model_family}, REAL consistently improves both veracity prediction and FAE scores across all datasets over the base model. These observed superior performances are in line with the Llama-3 family, confirming that REAL generalises across model architectures and pre-training corpora.

\vspace{-8pt}

\subsection{Analysis of Representation Shifts under Evidence Ablation (RQ4)}
\label{ssec:result_representation}

FAE results demonstrate that standard verifier predictions often persist despite the ablation of supporting evidence. To investigate whether this persistence is driven by a shift toward parametric decision-making, we analyse the verifier’s internal dynamics in the representation space. Following exploration in \S\ref{sec:ideal}, we further investigate these two fact-checking paradigms:

\noindent\textbf{\textit{Knowledge-based Fact-Checking.}}
In the knowledge-based paradigm, the verifier is provided only with the system instruction and claim $c$, without any documentary evidence, yielding a representation vector $\mathbf{h}^{(K)(c)}$ that relies solely on parametric knowledge. 

\noindent\textbf{\textit{Evidence-based Fact-Checking.}}
During the iterative evidence ablation in FAE \S\ref{ssec:ablation_process}, we extract a representation vector $\mathbf{h}^{(r)}(c, S^r)$ at each ablation round $r$ to track the evolution of the verifier's decision state.

These representations are obtained from the final-layer hidden state at the last input token, immediately before output generation (\textbf{Figure~\ref{fig:prompt_probe}}). This fixed probe position ensures a consistent measurement of the decision state as the availability of evidence decreases.

\noindent\textbf{\textit{Directional alignment to knowledge-based.}}

\noindent
To quantify the shift toward parametric decision-making during evidence ablation, we measure the directional alignment between the ablation-induced shift and the vector leading to the knowledge-based state \cite{li2016understanding,carlini2021extracting}. For each ablation round $r \ge 1$, we compute the cosine similarity between the current representational displacement and the vector directing from the evidence-based state to the knowledge-based state:
\[
  \mathrm{Dir}_{\mathrm{K}}^{(r)}
  = \cos\!\left(
      \mathbf{h}^{(r)}(c,S^r) - \mathbf{h}^{(0)}(c,S^0),\;
      \mathbf{h}^{(K)}(c) - \mathbf{h}^{(0)}(c,S^0)
    \right).
\]

This formulation isolates the direction of representational change induced by evidence ablation, rather than absolute similarity between states.
A higher $\mathrm{Dir}_{\mathrm{K}}^{(r)}$ indicates that ablating evidence causes the model's internal state to drift increasingly along the "evidence-to-knowledge" axis. This progressive drift signifies that as evidence becomes unavailable, the verifier systematically reverts to its parametric knowledge-based state to make a decision.

\begin{figure}[t]
\centering
\small
\setlength{\fboxsep}{4pt}
\fcolorbox{black}{gray!10}{
\parbox{0.9\linewidth}{
\textbf{Evidence-Based Fact-checking (with context).}\\
\texttt{[SYSTEM]} Verify the claim strictly against the provided evidence.\\
\texttt{[USER]} Claim: \emph{c}\\
Document:\\
\texttt{line12: ...}\\
\texttt{line27: ...}\\
\texttt{line45: ...}\\
\hfill $\uparrow$ \emph{probe position}
}
}

\vspace{4pt}
\fcolorbox{black}{gray!10}{
\parbox{0.9\linewidth}{
\textbf{Knowledge-Based Fact-checking (zero-context).}\\
\texttt{[SYSTEM]} Verify the claim using general knowledge.\\
\texttt{[USER]} Claim: \emph{c}\\
\hfill $\uparrow$ \emph{probe position}
}
}

\caption{Illustration of two paradigms. The probe extracts the final-layer hidden state at the last input token, immediately before output generation, under identical formatting conditions.}
\Description{Illustration of evidence-based and knowledge-based inputs. The probe extracts the final-layer hidden state at the last input token, immediately before output generation, under identical formatting conditions.}
\label{fig:prompt_probe}
\vspace{-10pt}
\end{figure}

\noindent\textbf{\textit{Results.}}
Figure~\ref{fig:dire_to_zero} reports the alignment trajectories of verifiers' hidden states relative to the knowledge-based paradigm. All trajectories and their 95\% bootstrap confidence intervals are computed over 2,000 iterations to ensure statistical robustness.

On FEVER, SciFact, and Check-COVID, the base verifier shows a sharp, monotonic increase in $\mathrm{Dir}_{\mathrm{K}}^{(r)}$ as evidence is ablated. This trend confirms that the base model's hidden states shift toward its parametric knowledge state. Notably, on FEVER and Check-COVID, the base model's confidence intervals become entirely disjoint from those of the REAL-trained model by the second ablation round. Even on SciFact, the base model’s mean alignment increases from 0.427 to 0.456, confirming a systematic reversion to parametric knowledge.

In contrast, REAL-trained verifiers maintain flatter trajectories across all benchmarks. Despite the loss of evidence, their hidden states remain distinct from the knowledge-based paradigm. This suppression of directional drift is clear on FEVER, where the model does not shift toward its parametric knowledge. Similar stability is observed in SciFact and Check-COVID, where the trained model maintains a stable mean alignment (e.g., approximately 0.430 on SciFact). These results suggest that REAL decouples the decision process from parametric knowledge, preventing the model from falling back on parametric knowledge when evidence is unavailable.

On Climate-FEVER, both models show higher proximity to the knowledge-based paradigm with overlapping confidence intervals. This reflects a stronger tendency of models to rely on parametric knowledge, likely due to the specific nature of this dataset. Unlike the other three datasets which provide coherent and complete documents, Climate-FEVER consists of isolated sentences from multiple sources, making it harder to form a robust evidentiary representation. Nevertheless, REAL still maintains a lower mean trajectory than the base model, showing persistent grounding pressure even in such fragmented, knowledge-intensive scenarios.

These representational probes reveal the mechanism behind the observed prediction persistence. REAL suppresses the drift toward parametric baselines, ensuring that the verifier’s hidden states remain anchored to the availability of external evidence.

\subsection{Prediction Stability}

\noindent In some rare cases, the accuracy drop does not correspond to the model switching its prediction to \texttt{NOT\_ENOUGH\_INFO} after evidence ablation but instead reflects unstable behaviour, in which the model changes its answers arbitrarily, in contrast to the ideal behaviour (\S\ref{sec:ideal}). For example, a model flipping its judgment from \texttt{SUPPORT} to \texttt{REFUTE} after an ablation round indicates guessing rather than evidence-based fact-checking. A model should ideally transition toward the neutral state (\texttt{NOT\_ENOUGH\_INFO}) when evidence is ablated, rather than switching to the opposite label.

To ensure FAE is reliably evaluated, we measure the frequency of contradictory label flips, where the prediction switches directly between \texttt{SUPPORT} and \texttt{REFUTE} during any ablation round. Table~\ref{tab:contradictory_flips} reports the average percentage of such flips. We observe that contradictory flips are remarkably rare, where the highest flip rate is only $2.98\%$, observed with Llama-3.1-8B-Instruct on the SciFact task. This low average frequency confirms that the performance decay observed in FAE is not an artefact of label instability. Instead, it demonstrates that models are consistently shifting toward a ``neutral'' stance as information is ablated, reinforcing the reliability of our ablation framework.

\begin{table}[t]
\centering
\caption{Percentage of SUPPORT--REFUTE label flips.}
\label{tab:contradictory_flips}
\resizebox{0.42\textwidth}{!}{
\begin{tabular}{lcccc}
\hline
\textbf{Mode}         & \textbf{FEVER} & \textbf{SciFact} & \textbf{\begin{tabular}[c]{@{}c@{}}Climate-\\ FEVER\end{tabular}} & \textbf{\begin{tabular}[c]{@{}c@{}}Check-\\ COVID\end{tabular}} \\ \hline
GPT-4o-mini           & 0.50           & 1.00             & 0.91                                                              & 0.79                                                            \\
Gemini-2.5-flash-lite & 0.16           & 0.72             & 0.41                                                              & 0.37                                                            \\
Qwen-2.5-32B-Instruct & 0.15           & 0.45             & 0.36                                                              & 0.59                                                            \\
Llama-3.1-8B-Instruct & 1.00           & 2.98             & 2.51                                                              & 2.22                                                            \\
REAL               & 0.05           & 0.42             & 0.22                                                              & 0.57                                                            \\ \hline
\end{tabular}
}
\vspace{-7pt}

\end{table}

Interestingly, we further observe that after applying REAL, the percentage of such label flips decreases (from $2.98\%$ to $0.42\%$) and this trend is consistent across all evaluation datasets. This observation further highlights the effectiveness of REAL in inducing truly evidence-grounded behaviour in fact-checking.

\section{Conclusion}

This paper investigates a critical limitation in LLM-based automated fact-checking that strong verification accuracy does not necessarily imply evidence-grounded reasoning. Through Fact Ablated Evaluation (FAE), we show that current LLM verifiers often preserve their predictions even after the supporting evidence has been removed, revealing substantial reliance on parametric knowledge during verification. To address this issue, we propose Rigorous Evidence Ablation Learning (REAL), a training framework that explicitly supervises verifier behaviour under both complete evidence and ablated evidence conditions through counterfactual evidence supervision and enhanced evidence annotation. Experiments across four fact-checking benchmarks demonstrate that REAL substantially improves evidence-grounded behaviour while maintaining strong fact-checking performance under both in-domain and out-of-domain evaluation. Our results further show that many existing verifiers can maintain correct predictions even after the supporting evidence has been removed, indicating that standard fact-checking accuracy alone may not fully reflect whether predictions are genuinely supported by the retrieved evidence. By explicitly evaluating and supervising verifier behaviour under evidence ablation, this work provides a practical framework for analysing and improving evidence dependency in LLM-based fact-checking systems.

\section*{Limitations}
FAE requires iterative evidence ablation and repeated verifier inference, making evaluation computationally more expensive than standard single-pass fact-checking evaluation. 
Consequently, FAE is currently suitable as a diagnostic framework for analysing evidence dependency rather than as a lightweight large-scale evaluation protocol. 
In addition, evidence redundancy and incomplete annotations in existing fact-checking benchmarks may still allow models to preserve correct predictions after `gold' evidence removal, making it difficult to perfectly separate evidence-grounded verification from parametric recall.
Although REAL partially mitigates this issue through evidence enhancement, complete isolation of all supporting evidence cannot always be guaranteed. 


\section*{GenAI Usage Disclosure}
This manuscript has benefited from the use of Generative AI (ChatGPT) to improve the quality of text produced by the authors. The authors remain responsible for the content.

\balance
\bibliographystyle{ACM-Reference-Format}
\bibliography{sample-base}

@inproceedings{hu2022lora,
  title={Lora: Low-rank adaptation of large language models.},
  author={Hu, Edward J and Shen, Yelong and Wallis, Phillip and Allen-Zhu, Zeyuan and Li, Yuanzhi and Wang, Shean and Wang, Lu and Chen, Weizhu and others},
  booktitle={Proc. of ICLR},
  year={2022}
}

@article{guo2022survey,
  title={A survey on automated fact-checking},
  author={Guo, Zhijiang and Schlichtkrull, Michael and Vlachos, Andreas},
  journal={Transactions of the Association for Computational Linguistics},
  volume={10},
  pages={178--206},
  year={2022}
}

@inproceedings{chenghaozhu2025your,
  title={Is Your LLM Outdated? A Deep Look at Temporal Generalization},
  author={ChenghaoZhu, ChenghaoZhu and Chen, Nuo and Gao, Yufei and Zhang, Yunyi and Tiwari, Prayag and Wang, Benyou},
  booktitle={Proc. of NAACL},
  pages={7433--7457},
  year={2025}
}

@article{das2025security,
  title={Security and privacy challenges of large language models: A survey},
  author={Das, Badhan Chandra and Amini, M Hadi and Wu, Yanzhao},
  journal={ACM Computing Surveys},
  volume={57},
  number={6},
  pages={1--39},
  year={2025}
}

@inproceedings{
song2025measuring,
title={Measuring and Enhancing Trustworthiness of {LLM}s in {RAG} through Grounded Attributions and Learning to Refuse},
author={Maojia Song and Shang Hong Sim and Rishabh Bhardwaj and Hai Leong Chieu and Navonil Majumder and Soujanya Poria},
booktitle={Proc. of ICLR},
year={2025}
}

@inproceedings{
joren2025sufficient,
title={Sufficient Context: A New Lens on Retrieval Augmented Generation Systems},
author={Hailey Joren and Jianyi Zhang and Chun-Sung Ferng and Da-Cheng Juan and Ankur Taly and Cyrus Rashtchian},
booktitle={Proc. of ICLR},
year={2025}
}

@inproceedings{niu-etal-2024-ragtruth,
    title = "{RAGT}ruth: A Hallucination Corpus for Developing Trustworthy Retrieval-Augmented Language Models",
    author = "Niu, Cheng  and
      Wu, Yuanhao  and
      Zhu, Juno  and
      Xu, Siliang  and
      Shum, KaShun  and
      Zhong, Randy  and
      Song, Juntong  and
      Zhang, Tong",
    booktitle = "Proc of ACL",
    year = "2024",
}

@inproceedings{ru2024ragchecker,
  title={RAGCHECKER: a fine-grained framework for diagnosing retrieval-augmented generation},
  author={Ru, Dongyu and Qiu, Lin and Hu, Xiangkun and Zhang, Tianhang and Shi, Peng and Chang, Shuaichen and Jiayang, Cheng and Wang, Cunxiang and Sun, Shichao and Li, Huanyu and others},
  booktitle={Proc. of NeurIPS},
  year={2024}
}

@inproceedings{lewis2020retrieval,
  title={Retrieval-augmented generation for knowledge-intensive nlp tasks},
  author={Lewis, Patrick and Perez, Ethan and Piktus, Aleksandra and Petroni, Fabio and Karpukhin, Vladimir and Goyal, Naman and K{\"u}ttler, Heinrich and Lewis, Mike and Yih, Wen-tau and Rockt{\"a}schel, Tim and others},
  booktitle={Proc. of NeurIPS},
  year={2020}
}

@inproceedings{thorne-etal-2018-fever,
    title = "{FEVER}: a Large-scale Dataset for Fact Extraction and {VER}ification",
    author = "Thorne, James  and
      Vlachos, Andreas  and
      Christodoulopoulos, Christos  and
      Mittal, Arpit",
    booktitle = "Proc. of NAACL",
    year = "2018"
}

@inproceedings{thakur-etal-2024-knowing,
    title = "``Knowing When You Don{'}t Know'': A Multilingual Relevance Assessment Dataset for Robust Retrieval-Augmented Generation",
    author = "Thakur, Nandan  and
      Bonifacio, Luiz  and
      Zhang, Crystina  and
      Ogundepo, Odunayo  and
      Kamalloo, Ehsan  and
      Alfonso-Hermelo, David  and
      Li, Xiaoguang  and
      Liu, Qun  and
      Chen, Boxing  and
      Rezagholizadeh, Mehdi  and
      Lin, Jimmy",
    booktitle = "Findings of the Association for Computational Linguistics: EMNLP 2024",
    year = "2024"
}

@inproceedings{ye-etal-2024-effective,
    title = "Effective Large Language Model Adaptation for Improved Grounding and Citation Generation",
    author = "Ye, Xi  and
      Sun, Ruoxi  and
      Arik, Sercan  and
      Pfister, Tomas",
    booktitle = "Proc. of NAACL",
    year = "2024",
}

@inproceedings{hsu-etal-2024-calm,
    title = "{C}a{LM}: Contrasting Large and Small Language Models to Verify Grounded Generation",
    author = "Hsu, I-Hung  and
      Wang, Zifeng  and
      Le, Long  and
      Miculicich, Lesly  and
      Peng, Nanyun  and
      Lee, Chen-Yu  and
      Pfister, Tomas",
    booktitle = "Findings of the Association for Computational Linguistics: ACL 2024",
    year = "2024",
}

@inproceedings{huang-etal-2024-learning,
    title = "Learning Fine-Grained Grounded Citations for Attributed Large Language Models",
    author = "Huang, Lei  and
      Feng, Xiaocheng  and
      Ma, Weitao  and
      Gu, Yuxuan  and
      Zhong, Weihong  and
      Feng, Xiachong  and
      Yu, Weijiang  and
      Peng, Weihua  and
      Tang, Duyu  and
      Tu, Dandan  and
      Qin, Bing",
    booktitle = "Findings of the Association for Computational Linguistics: ACL 2024",
    year = "2024",
}

@article{guo-etal-2022-survey,
    title = "A Survey on Automated Fact-Checking",
    author = "Guo, Zhijiang  and
      Schlichtkrull, Michael  and
      Vlachos, Andreas",
    journal = "Transactions of the Association for Computational Linguistics",
    volume = "10",
    year = "2022",
}

@article{zeng2021automated,
  title={Automated fact-checking: A survey},
  author={Zeng, Xia and Abumansour, Amani S and Zubiaga, Arkaitz},
  journal={Language and Linguistics Compass},
  volume={15},
  number={10},
  pages={e12438},
  year={2021}
}

@inproceedings{vladika-matthes-2023-scientific,
    title = "Scientific Fact-Checking: A Survey of Resources and Approaches",
    author = "Vladika, Juraj  and
      Matthes, Florian",
    booktitle = "Findings of the Association for Computational Linguistics: ACL 2023",
    year = "2023",
}

@inproceedings{tang-etal-2024-minicheck,
    title = "{M}ini{C}heck: Efficient Fact-Checking of {LLM}s on Grounding Documents",
    author = "Tang, Liyan  and
      Laban, Philippe  and
      Durrett, Greg",
    booktitle = "Proc. of EMNLP",
    year = "2024",
}

@inproceedings{pan-etal-2023-fact,
    title = "Fact-Checking Complex Claims with Program-Guided Reasoning",
    author = "Pan, Liangming  and
      Wu, Xiaobao  and
      Lu, Xinyuan  and
      Luu, Anh Tuan  and
      Wang, William Yang  and
      Kan, Min-Yen  and
      Nakov, Preslav",
    booktitle = "Proc. of ACL",
    year = "2023",
}

@inproceedings{braun2025defame,
    title={{DEFAME}: Dynamic Evidence-based {FA}ct-checking with Multimodal Experts},
    author={Tobias Braun and Mark Rothermel and Marcus Rohrbach and Anna Rohrbach},
    booktitle={Proc. of ICML},
    year={2025},
}

@inproceedings{schlichtkrull-etal-2024-automated,
    title = "The Automated Verification of Textual Claims ({AV}eri{T}e{C}) Shared Task",
    author = "Schlichtkrull, Michael  and
      Chen, Yulong  and
      Whitehouse, Chenxi  and
      Deng, Zhenyun  and
      Akhtar, Mubashara  and
      Aly, Rami  and
      Guo, Zhijiang  and
      Christodoulopoulos, Christos  and
      Cocarascu, Oana  and
      Mittal, Arpit  and
      Thorne, James  and
      Vlachos, Andreas",
    booktitle = "Proc. of FEVER",
    year = "2024",
}

@inproceedings{wadden-etal-2020-fact,
    title = "Fact or Fiction: Verifying Scientific Claims",
    author = "Wadden, David  and
      Lin, Shanchuan  and
      Lo, Kyle  and
      Wang, Lucy Lu  and
      van Zuylen, Madeleine  and
      Cohan, Arman  and
      Hajishirzi, Hannaneh",
    booktitle = "Proc. of EMNLP",
    year = "2020",
}

@article{diggelmann2020climate,
  title={Climate-fever: A dataset for verification of real-world climate claims},
  author={Diggelmann, Thomas and Boyd-Graber, Jordan and Bulian, Jannis and Ciaramita, Massimiliano and Leippold, Markus},
  journal={arXiv preprint arXiv:2012.00614},
  year={2020}
}

@inproceedings{wang-etal-2023-check-covid,
    title = "Check-{COVID}: Fact-Checking {COVID}-19 News Claims with Scientific Evidence",
    author = "Wang, Gengyu  and
      Harwood, Kate  and
      Chillrud, Lawrence  and
      Ananthram, Amith  and
      Subbiah, Melanie  and
      McKeown, Kathleen",
    booktitle = "Findings of the Association for Computational Linguistics: ACL 2023",
    year = "2023",
}

@inproceedings{derczynski-etal-2020-maintaining,
    title = "Maintaining Quality in {FEVER} Annotation",
    author = "Derczynski, Leon  and
      Binau, Julie  and
      Schulte, Henri",
    booktitle = "Proc. of FEVER",
    year = "2020",
}

@inproceedings{vladika-etal-2025-facts,
    title = "Facts Fade Fast: Evaluating Memorization of Outdated Medical Knowledge in Large Language Models",
    author = "Vladika, Juraj  and
      Dhaini, Mahdi  and
      Matthes, Florian",
    booktitle = "Findings of the Association for Computational Linguistics: EMNLP 2025",
    year = "2025",
}

@inproceedings{vlachos-riedel-2014-fact,
    title = "Fact Checking: Task definition and dataset construction",
    author = "Vlachos, Andreas  and
      Riedel, Sebastian",
    booktitle = "Proc. of ACL",
    year = "2014",
}

@inproceedings{akhtar-etal-2025-2nd,
    title = "The 2nd Automated Verification of Textual Claims ({AV}eri{T}e{C}) Shared Task: Open-weights, Reproducible and Efficient Systems",
    author = "Akhtar, Mubashara  and
      Aly, Rami  and
      Chen, Yulong  and
      Deng, Zhenyun  and
      Schlichtkrull, Michael  and
      Whitehouse, Chenxi  and
      Vlachos, Andreas",
    booktitle = "Proc. of FEVER",
    year = "2025",
}

@inproceedings{kassem-etal-2025-alpaca,
    title = "{ALPACA} {AGAINST} {VICUNA}: Using {LLM}s to Uncover Memorization of {LLM}s",
    author = "Kassem, Aly M.  and
      Mahmoud, Omar  and
      Mireshghallah, Niloofar  and
      Kim, Hyunwoo  and
      Tsvetkov, Yulia  and
      Choi, Yejin  and
      Saad, Sherif  and
      Rana, Santu",
    booktitle = "Proc. of NAACL",
    year = "2025",
}

@inproceedings{
    schlichtkrull2023averitec,
    title={{AV}eriTeC: A Dataset for Real-world Claim Verification with Evidence from the Web},
    author={Michael Sejr Schlichtkrull and Zhijiang Guo and Andreas Vlachos},
    booktitle={Thirty-seventh Conference on Neural Information Processing Systems Datasets and Benchmarks Track},
    year={2023},
}

@inproceedings{hu2023read,
  title={Read it twice: Towards faithfully interpretable fact verification by revisiting evidence},
  author={Hu, Xuming and Hong, Zhaochen and Guo, Zhijiang and Wen, Lijie and Yu, Philip},
  booktitle={Proc. of SIGIR}, 
  year={2023}
}

@inproceedings{zheng-etal-2024-evidence,
    title = "Evidence Retrieval is almost All You Need for Fact Verification",
    author = "Zheng, Liwen  and
      Li, Chaozhuo  and
      Zhang, Xi  and
      Shang, Yu-Ming  and
      Huang, Feiran  and
      Jia, Haoran",
    booktitle = "Findings of the Association for Computational Linguistics: ACL 2024",
    year = "2024",
}

@inproceedings{vladika-matthes-2024-improving,
    title = "Improving Health Question Answering with Reliable and Time-Aware Evidence Retrieval",
    author = "Vladika, Juraj  and
      Matthes, Florian",
    booktitle = "Findings of the Association for Computational Linguistics: NAACL 2024",
    year = "2024",
}

@inproceedings{honovich-etal-2022-true-evaluating,
    title = "{TRUE}: Re-evaluating Factual Consistency Evaluation",
    author = "Honovich, Or  and
      Aharoni, Roee  and
      Herzig, Jonathan  and
      Taitelbaum, Hagai  and
      Kukliansy, Doron  and
      Cohen, Vered  and
      Scialom, Thomas  and
      Szpektor, Idan  and
      Hassidim, Avinatan  and
      Matias, Yossi",
    booktitle = "Proc. of NAACL",
    year = "2022",
}

@inproceedings{tang-etal-2025-enhancing,
    title = "Enhancing Chain-of-Thought Reasoning via Neuron Activation Differential Analysis",
    author = "Tang, Yiru  and
      Zhou, Kun  and
      Min, Yingqian  and
      Zhao, Wayne Xin  and
      Sha, Jing  and
      Sheng, Zhichao  and
      Wang, Shijin",
    booktitle = "Proc. of EMNLP",
    year = "2025",
}

@article{bekoulis2021review,
  title={A review on fact extraction and verification},
  author={Bekoulis, Giannis and Papagiannopoulou, Christina and Deligiannis, Nikos},
  journal={ACM Computing Surveys (CSUR)},
  volume={55},
  number={1},
  pages={1--35},
  year={2021}, 
}

@inproceedings{hanselowski-etal-2019-richly,
    title = "A Richly Annotated Corpus for Different Tasks in Automated Fact-Checking",
    author = "Hanselowski, Andreas  and
      Stab, Christian  and
      Schulz, Claudia  and
      Li, Zile  and
      Gurevych, Iryna",
    booktitle = "Proc. of CoNLL",
    year = "2019",
}

@inproceedings{wu2021evidence,
  title={Evidence-aware hierarchical interactive attention networks for explainable claim verification},
  author={Wu, Lianwei and Rao, Yuan and Yang, Xiong and Wang, Wanzhen and Nazir, Ambreen},
  booktitle={Proc. of IJCAI}, 
  year={2021}
}

@inproceedings{chen2022gere,
  title={GERE: Generative evidence retrieval for fact verification},
  author={Chen, Jiangui and Zhang, Ruqing and Guo, Jiafeng and Fan, Yixing and Cheng, Xueqi},
  booktitle={Proc. of SIGIR},
  year={2022}
}

@inproceedings{luken-etal-2018-qed,
    title = "{QED}: A fact verification system for the {FEVER} shared task",
    author = "Luken, Jackson  and
      Jiang, Nanjiang  and
      de Marneffe, Marie-Catherine",
    booktitle = "Proc. of FEVER",
    year = "2018",
}

@inproceedings{nie2019combining,
  title={Combining fact extraction and verification with neural semantic matching networks},
  author={Nie, Yixin and Chen, Haonan and Bansal, Mohit},
  booktitle={Proc. of AAAI},
  year={2019}
}

@misc{qwen2025qwen25technicalreport,
      title={Qwen2.5 Technical Report}, 
      author={Qwen and : and An Yang and Baosong Yang and Beichen Zhang and Binyuan Hui and Bo Zheng and Bowen Yu and Chengyuan Li and Dayiheng Liu and Fei Huang and Haoran Wei and Huan Lin and Jian Yang and Jianhong Tu and Jianwei Zhang and Jianxin Yang and Jiaxi Yang and Jingren Zhou and Junyang Lin and Kai Dang and Keming Lu and Keqin Bao and Kexin Yang and Le Yu and Mei Li and Mingfeng Xue and Pei Zhang and Qin Zhu and Rui Men and Runji Lin and Tianhao Li and Tianyi Tang and Tingyu Xia and Xingzhang Ren and Xuancheng Ren and Yang Fan and Yang Su and Yichang Zhang and Yu Wan and Yuqiong Liu and Zeyu Cui and Zhenru Zhang and Zihan Qiu},
      year={2025},
      eprint={2412.15115},
      archivePrefix={arXiv},
      primaryClass={cs.CL},
      url={https://arxiv.org/abs/2412.15115}, 
}

@article{comanici2025gemini,
  title={Gemini 2.5: Pushing the frontier with advanced reasoning, multimodality, long context, and next generation agentic capabilities},
  author={Comanici, Gheorghe and Bieber, Eric and Schaekermann, Mike and Pasupat, Ice and Sachdeva, Noveen and Dhillon, Inderjit and Blistein, Marcel and Ram, Ori and Zhang, Dan and Rosen, Evan and others},
  journal={arXiv preprint arXiv:2507.06261},
  year={2025}
}

@article{hurst2024gpt,
  title={Gpt-4o system card},
  author={Hurst, Aaron and Lerer, Adam and Goucher, Adam P and Perelman, Adam and Ramesh, Aditya and Clark, Aidan and Ostrow, AJ and Welihinda, Akila and Hayes, Alan and Radford, Alec and others},
  journal={arXiv preprint arXiv:2410.21276},
  year={2024}
}

@article{grattafiori2024llama,
  title={The llama 3 herd of models},
  author={Grattafiori, Aaron and Dubey, Abhimanyu and Jauhri, Abhinav and Pandey, Abhinav and Kadian, Abhishek and Al-Dahle, Ahmad and Letman, Aiesha and Mathur, Akhil and Schelten, Alan and Vaughan, Alex and others},
  journal={arXiv preprint arXiv:2407.21783},
  year={2024}
}

@inproceedings{wadden-lo-2021-overview,
    title = "Overview and Insights from the {SCIVER} shared task on Scientific Claim Verification",
    author = "Wadden, David  and
      Lo, Kyle",
    booktitle = "Proceedings of the Second Workshop on Scholarly Document Processing",
    year = "2021",
}

@article{ouyang2022training,
  title={Training language models to follow instructions with human feedback},
  author={Ouyang, Long and Wu, Jeffrey and Jiang, Xu and Almeida, Diogo and Wainwright, Carroll and Mishkin, Pamela and Zhang, Chong and Agarwal, Sandhini and Slama, Katarina and Ray, Alex and others},
  journal={Advances in neural information processing systems},
  volume={35},
  pages={27730--27744},
  year={2022}
}

@article{gao2023retrieval,
  title={Retrieval-augmented generation for large language models: A survey},
  author={Gao, Yunfan and Xiong, Yun and Gao, Xinyu and Jia, Kangxiang and Pan, Jinliu and Bi, Yuxi and Dai, Yixin and Sun, Jiawei and Wang, Haofen and Wang, Haofen},
  journal={arXiv preprint arXiv:2312.10997},
  volume={2},
  number={1},
  year={2023}
}

@inproceedings{yang2025knowing,
  title={Knowing You Don't Know: Learning When to Continue Search in Multi-round RAG through Self-Practicing},
  author={Yang, Diji and Zeng, Linda and Rao, Jinmeng and Zhang, Yi},
  booktitle={Proc. of SIGIR},
  year={2025}
}

@article{akhtar2024ev2r,
  title={Ev2r: Evaluating evidence retrieval in automated fact-checking},
  author={Akhtar, Mubashara and Schlichtkrull, Michael and Vlachos, Andreas},
  journal={arXiv preprint arXiv:2411.05375},
  year={2024}
}

@inproceedings{
loshchilov2018decoupled,
title={Decoupled Weight Decay Regularization},
author={Ilya Loshchilov and Frank Hutter},
booktitle={International Conference on Learning Representations},
year={2019},
url={https://openreview.net/forum?id=Bkg6RiCqY7},
}

@inproceedings{rothermel-etal-2024-infact,
    title = "{I}n{F}act: A Strong Baseline for Automated Fact-Checking",
    author = "Rothermel, Mark  and
      Braun, Tobias  and
      Rohrbach, Marcus  and
      Rohrbach, Anna",
    booktitle = "Proc. of FEVER",
    year = "2024",
}

@inproceedings{yoon-etal-2024-hero,
    title = "{H}er{O} at {AV}eri{T}e{C}: The Herd of Open Large Language Models for Verifying Real-World Claims",
    author = "Yoon, Yejun  and
      Jung, Jaeyoon  and
      Yoon, Seunghyun  and
      Park, Kunwoo",
    booktitle = "Proc. of FEVER",
    year = "2024",
}

@inproceedings{yoon-etal-2025-team,
    title = "Team {HUMANE} at {AV}eri{T}e{C} 2025: {H}er{O} 2 for Efficient Fact Verification",
    author = "Yoon, Yejun  and
      Jung, Jaeyoon  and
      Yoon, Seunghyun  and
      Park, Kunwoo",
    booktitle = "Proc. of FEVER",
    year = "2025",
}

@inproceedings{ullrich-etal-2024-aic,
    title = "{AIC} {CTU} system at {AV}eri{T}e{C}: Re-framing automated fact-checking as a simple {RAG} task",
    author = "Ullrich, Herbert  and
      Mlyn{\'a}{\v{r}}, Tom{\'a}{\v{s}}  and
      Drchal, Jan",
    booktitle = "Proc. of FEVER",
    year = "2024",
}

@inproceedings{ullrich-drchal-2025-aic,
    title = "{AIC} {CTU}@{FEVER} 8: On-premise fact checking through long context {RAG}",
    author = "Ullrich, Herbert  and
      Drchal, Jan",
    booktitle = "Proc. of FEVER",
    month = jul,
    year = "2025",
}

@inproceedings{park-etal-2024-dunamu,
    title = "Dunamu-ml{'}s Submissions on {AVERITEC} Shared Task",
    author = "Park, Heesoo  and
      Lee, Dongjun  and
      Kim, Jaehyuk  and
      Park, ChoongWon  and
      Park, Changhwa",
    booktitle = "Proc. of FEVER",
    year = "2024",
}

@inproceedings{mohammadkhani-etal-2024-zero,
    title = "Zero-Shot Learning and Key Points Are All You Need for Automated Fact-Checking",
    author = "Mohammadkhani, Mohammad Ghiasvand  and
      Mohammadkhani, Ali Ghiasvand  and
      Beigy, Hamid",
    booktitle = "Proc. of FEVER",
    year = "2024",
}

@inproceedings{jannah-etal-2025-multilingual,
    title = "Multilingual Symptom Detection on Social Media: Enhancing Health-related Fact-checking with {LLM}s",
    author = "Jannah, Saidah Zahrotul  and
      Aco, Elyanah  and
      Peng, Shaowen  and
      Wakamiya, Shoko  and
      Aramaki, Eiji",
    booktitle = "Proc. of FEVER",
    month = jul,
    year = "2025",
}

@inproceedings{shcharbakova-etal-2025-scale,
    title = "When Scale Meets Diversity: Evaluating Language Models on Fine-Grained Multilingual Claim Verification",
    author = "Shcharbakova, Hanna  and
      Anikina, Tatiana  and
      Skachkova, Natalia  and
      Genabith, Josef Van",
    booktitle = "Proc. of FEVER",
    year = "2025",
}

@inproceedings{chowdhury-etal-2025-fact5,
    title = "{FACT}5: A Novel Benchmark and Pipeline for Nuanced Fact-Checking of Complex Statements",
    author = "Chowdhury, Shayan  and
      Fang, Sunny  and
      Muresan, Smaranda",
    booktitle = "Proc. of FEVER",
    year = "2025",
}

@article{li2025use,
  title={Use of Retrieval-Augmented Large Language Model for COVID-19 Fact-Checking: Development and Usability Study},
  author={Li, Hai and Huang, Jingyi and Ji, Mengmeng and Yang, Yuyi and An, Ruopeng},
  journal={Journal of medical Internet research},
  volume={27},
  pages={e66098},
  year={2025},
}

@inproceedings{zhou-etal-2025-gqc,
    title = "{GQC}: {LLM}-Based Grouped {QA} Consolidation for Open-Domain Fact Verification at {AV}eri{T}e{C}",
    author = "Zhou, Dongzhuoran  and
      Pop, Roxana  and
      Zhu, Yuqicheng  and
      Kharlamov, Evgeny",
    booktitle = "Proc. of FEVER",
    year = "2025",
}

@inproceedings{wang-etal-2024-factcheck,
    title = "Factcheck-Bench: Fine-Grained Evaluation Benchmark for Automatic Fact-checkers",
    author = "Wang, Yuxia  and
      Gangi Reddy, Revanth  and
      Mujahid, Zain Muhammad  and
      Arora, Arnav  and
      Rubashevskii, Aleksandr  and
      Geng, Jiahui  and
      Mohammed Afzal, Osama  and
      Pan, Liangming  and
      Borenstein, Nadav  and
      Pillai, Aditya  and
      Augenstein, Isabelle  and
      Gurevych, Iryna  and
      Nakov, Preslav",
    booktitle = "Findings of the Association for Computational Linguistics: EMNLP 2024",
    year = "2024",
}

@inproceedings{xie-etal-2025-fire,
    title = "{FIRE}: Fact-checking with Iterative Retrieval and Verification",
    author = "Xie, Zhuohan  and
      Xing, Rui  and
      Wang, Yuxia  and
      Geng, Jiahui  and
      Iqbal, Hasan  and
      Sahnan, Dhruv  and
      Gurevych, Iryna  and
      Nakov, Preslav",
    booktitle = "Findings of the Association for Computational Linguistics: NAACL 2025",
    year = "2025",
}

@inproceedings{
wei2024longform,
title={Long-form factuality in large language models},
author={Jerry Wei and Chengrun Yang and Xinying Song and Yifeng Lu and Nathan Zixia Hu and Jie Huang and Dustin Tran and Daiyi Peng and Ruibo Liu and Da Huang and Cosmo Du and Quoc V Le},
booktitle={The Thirty-eighth Annual Conference on Neural Information Processing Systems},
year={2024},
}

@inproceedings{cheung2023factllama,
  title={Factllama: Optimizing instruction-following language models with external knowledge for automated fact-checking},
  author={Cheung, Tsun-Hin and Lam, Kin-Man},
  booktitle={2023 Asia Pacific Signal and Information Processing Association Annual Summit and Conference (APSIPA ASC)},
  pages={846--853},
  year={2023},
}

@article{kumar2025improving,
  title={Improving the fact-checking performance of language models by relying on their entailment ability},
  author={Kumar, Gaurav and Mazumder, Debajyoti and Garg, Ayush and Patro, Jasabanta},
  journal={arXiv preprint arXiv:2505.15050},
  year={2025}
}

@article{geng2025m4fc,
  title={M4FC: a Multimodal, Multilingual, Multicultural, Multitask Real-World Fact-Checking Dataset},
  author={Geng, Jiahui and Tonglet, Jonathan and Gurevych, Iryna},
  journal={arXiv preprint arXiv:2510.23508},
  year={2025}
}

@inproceedings{putta-etal-2025-claimcheck,
    title = "{C}laim{C}heck: Automatic Fact-Checking of Textual Claims using Web Evidence",
    author = "Putta, Akshith Reddy  and
      Devasier, Jacob  and
      Li, Chengkai",
    booktitle = "Proceedings of the 4th International Workshop on Knowledge-Augmented Methods for Natural Language Processing", 
    year = "2025",
}

@inproceedings{kiyomaru-etal-2024-comprehensive-analysis,
    title = "A Comprehensive Analysis of Memorization in Large Language Models",
    author = "Kiyomaru, Hirokazu  and
      Sugiura, Issa  and
      Kawahara, Daisuke  and
      Kurohashi, Sadao",
    booktitle = "Proceedings of the 17th International Natural Language Generation Conference",
    month = sep,
    year = "2024",
}

@article{wei2025memorization,
  title={Memorization in deep learning: A survey},
  author={Wei, Jiaheng and Zhang, Yanjun and Zhang, Leo Yu and Ding, Ming and Chen, Chao and Ong, Kok-Leong and Zhang, Jun and Xiang, Yang},
  journal={ACM Computing Surveys},
  volume={58},
  number={4},
  pages={1--35},
  year={2025},
  publisher={ACM New York, NY}
}

@inproceedings{di2025llms,
  title={Do llms memorize recommendation datasets? a preliminary study on movielens-1m},
  author={Di Palma, Dario and Merra, Felice Antonio and Sfilio, Maurizio and Anelli, Vito Walter and Narducci, Fedelucio and Di Noia, Tommaso},
  booktitle={Proceedings of the 48th International ACM SIGIR Conference on Research and Development in Information Retrieval},
  year={2025}
}

@article{li2016understanding,
  title={Understanding neural networks through representation erasure},
  author={Li, Jiwei and Monroe, Will and Jurafsky, Dan},
  journal={arXiv preprint arXiv:1612.08220},
  year={2016}
}

@inproceedings{carlini2021extracting,
  title={Extracting training data from large language models},
  author={Carlini, Nicholas and Tramer, Florian and Wallace, Eric and Jagielski, Matthew and Herbert-Voss, Ariel and Lee, Katherine and Roberts, Adam and Brown, Tom and Song, Dawn and Erlingsson, Ulfar and others},
  booktitle={30th USENIX security symposium (USENIX Security 21)},
  pages={2633--2650},
  year={2021}
}

@inproceedings{deng2025+,
  title={+ VeriRel: Verification Feedback to Enhance Document Retrieval for Scientific Fact Checking},
  author={Deng, Xingyu and Wang, Xi and Stevenson, Mark},
  booktitle={Proceedings of the 34th ACM International Conference on Information and Knowledge Management},
  pages={4706--4711},
  year={2025}
}

@inproceedings{deng2025next,
  title={The next phase of scientific fact-checking: advanced evidence retrieval from complex structured academic papers},
  author={Deng, Xingyu and Wang, Xi and Stevenson, Mark},
  booktitle={Proceedings of the 2025 International ACM SIGIR Conference on Innovative Concepts and Theories in Information Retrieval (ICTIR)},
  pages={436--448},
  year={2025}
}

@inproceedings{deng2026towards,
  title={Towards Evidence-Aware Retrieval and Verification for Scientific Fact-Checking},
  author={Deng, Xingyu},
  booktitle={Proceedings of the 49th International ACM SIGIR Conference on Research and Development in Information Retrieval},
  pages={5309--5309},
  year={2026}
}

\appendix

\end{document}